\pdfoutput=1
\documentclass[letterpaper, 10 pt, journal, twoside]{IEEEtran}
\usepackage{cite}
\usepackage{url}
\usepackage{amsmath,amssymb,amsfonts}
\usepackage[noend]{algpseudocode}
\usepackage{graphicx}
\usepackage{textcomp}
\usepackage{algorithm} 
\usepackage{float}
\usepackage{acronym}
\usepackage{booktabs}
\usepackage{siunitx}
\usepackage{threeparttable}
\usepackage{multirow}
\usepackage{makecell}
\usepackage[export]{adjustbox}
\usepackage{bm}

\usepackage{pifont} 

\usepackage[dvipsnames]{xcolor}
\DeclareSIUnit\px{px}

\usepackage{soul} 
\colorlet{soulyellow}{yellow!50}
\sethlcolor{soulyellow}

\acrodef{NeRFs}{Neural Radiance Fields}
\acrodef{SDFs}{Signed Distance Fields}
\acrodef{std}{standard deviation}
\acrodef{GANs}{Generative Adversarial Networks}
\acrodef{MAE}{mean absolute error}
\acrodef{MGD}{mean gradient difference}
\acrodef{PSNR}{peak signal-to-noise ratio}
\acrodef{SSIM}{structural similarity index measure}
\acrodef{NDT}{Normal Distribution Transform}
\acrodef{ICP}{Iterative Closest Point}

\acrodef{FMM}{Fast Marching Method}
\acrodef{ESDF}{Euclidean Signed Distance Field}
\acrodef{ESDFs}{Euclidean Signed Distance Fields}
\acrodef{TSDF}{Truncated Signed Distance Field}
\acrodef{TSDFs}{Truncated Signed Distance Fields}
\acrodef{CVP}{Continuous Vector-field planner}

\makeatletter
\let\NAT@parse\undefined
\makeatother
\usepackage[hidelinks]{hyperref} 

\begin{document}

\title{Efficient Global Navigational Planning in 3D Structures based on Point Cloud Tomography}
\author{Bowen Yang$^{1}$, Jie Cheng$^{1}$, Bohuan Xue$^{1}$, \textit{Graduate Student Member, IEEE}, \\ Jianhao Jiao$^{1, 2}$, \textit{Member, IEEE}, and Ming Liu$^{3}$, \textit{Senior Member, IEEE} %
\thanks{$^{1}$B. Yang, J. Cheng, B. Xue, and J. Jiao are with the Hong Kong University of Science and Technology, Hong Kong SAR, China. {\tt\small \{byangar,jchengai,bxueaa,jjiao\}@connect.ust.hk} \textit{(Corresponding author: Jianhao Jiao)}.}
\thanks{$^{2}$J. Jiao is also with the Department of Computer Science, University College London, Gower Street, WC1E 6BT, London, UK.}
\thanks{$^{3}$M. Liu is with the Hong Kong University of Science and Technology (Guangzhou), Nansha, Guangzhou, 511400, Guangdong, China. {\tt\small eelium@hkust-gz.edu.cn}}
}

\maketitle

\begin{abstract}
Navigation in complex 3D scenarios requires appropriate environment representation for efficient scene understanding and trajectory generation. 
We propose a highly efficient and extensible global navigation framework based on a tomographic understanding of the environment to navigate ground robots in multi-layer structures.
Our approach generates tomogram slices using the point cloud map to encode the geometric structure as ground and ceiling elevations.
Then it evaluates the scene traversability considering the robot's motion capabilities.
Both the tomogram construction and the scene evaluation are accelerated through parallel computation.
Our approach further alleviates the trajectory generation complexity compared with planning in 3D spaces directly. 
It generates 3D trajectories by searching through multiple tomogram slices and separately adjusts the robot height to avoid overhangs.
We evaluate our framework in various simulation scenarios and further test it in the real world on a quadrupedal robot. 
Our approach reduces the scene evaluation time by 3 orders of magnitude and improves the path planning speed by 3 times compared with existing approaches, demonstrating highly efficient global navigation in various complex 3D environments.
The code is available at: \href{https://github.com/byangw/PCT_planner}{https://github.com/byangw/PCT\_planner}.
\end{abstract}


\section{Introduction}
\IEEEPARstart{N}{avigating} ground robots in 3D environments is essential for a wide range of autonomous applications.
However, it's still challenging to efficiently evaluate the multi-layer scenarios with complex terrain conditions and spatial structures, where a proper environmental representation would assist to improve the scene evaluation speed.

Point clouds and meshes can represent detailed 3D structures and are applied in robot navigation problems \cite{roboticOnline,beyond2D}, while their irregular data structures may bring difficulties to scene understanding.
Voxels discretize the space into structured 3D grids for the convenience of construction and map processing, which are widely adopted to navigate UAVs \cite{zhouUAV,raptor}.
However, it's usually difficult to achieve high mapping efficiency while maintaining the capabilities to represent detailed environment structures.
In addition, unlike drones that fly in 3D space, voxels might not be the optimal choice for ground robots that are more concerned about terrain conditions.

\begin{figure}[t]
\setlength{\abovecaptionskip}{0pt}
\centering
\includegraphics[width=3.4in]{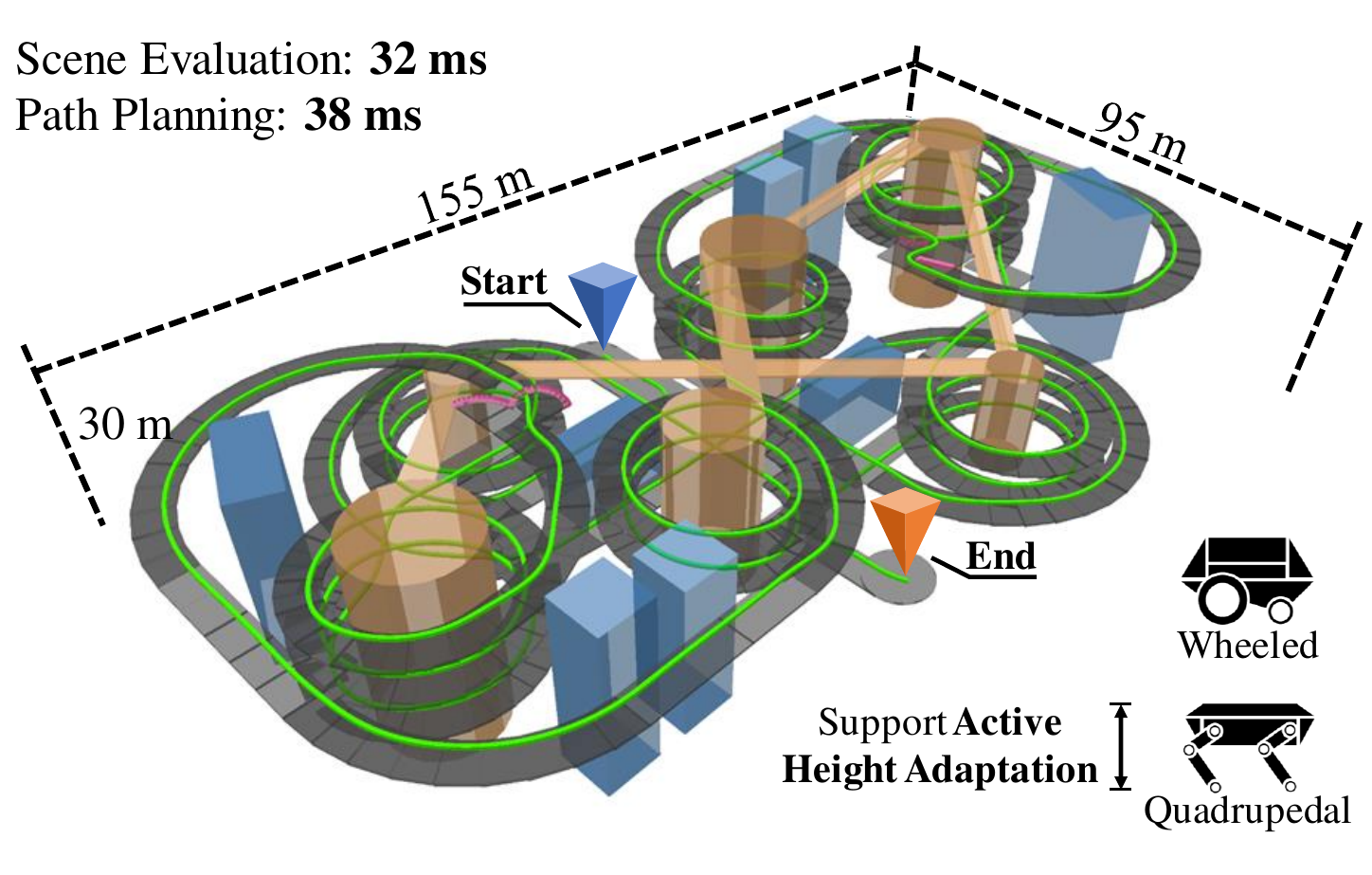}
\caption{
We present a highly efficient and extensible global navigation framework that generates smooth 3D trajectories in complex multi-layer scenarios.
It adopts a novel scene representation which enables rapid scene evaluation and further alleviates the burden of path searching. 
The framework applies to a wide range of ground robots including wheeled or quadrupedal robots and supports active height adaptation of the platform to avoid overhangs.
}
\label{fig:cover}
\vspace{-1.2em}
\end{figure}

Elevation maps \cite{elevation2014,elevation2018} are widely adopted in navigating ground robots on complex uneven terrains.
Compared with voxel-based representations, elevation maps store continuous ground height values in 2D grids to represent the terrain surface, which balances the mapping efficiency and the capability to represent detailed ground conditions. 
However, traditional elevation maps fail to identify overhangs or multi-layer structures, restricting their scope of application.
Triebel \textit{et al.} \cite{MLSmap} extend elevation maps by storing multi-level surfaces into grid-wise lists, which brings difficulties in scene evaluation.
Miki \textit{et al.} \cite{elevationCupy} adopt exclusion areas to reject overhangs and apply overlap clearance to update the ground levels in multi-layer structures, restricting it to only represent a single local terrain surface and the method is unsuitable for large-scale scenarios. 

We present a global navigation framework for ground robots in complex 3D scenarios with the following aspects contributing to its high efficiency.
First, we propose a novel scene representation approach based on a tomographic understanding of the point cloud.
It encodes the geometric structure into multiple tomogram slices containing the ground and ceiling elevations.
Our approach maintains the mapping simplicity and terrain representation capability of elevation maps while extending their scope to large-scale multi-layer scenarios.
In addition, our scene representation is compatible with a wide range of mapping, scene evaluation, and path planning methods on grid maps, making our framework highly extensible to further enhancements.
Second, a kernel-based scene evaluation method is designed for rapid traversability estimation on tomograms.
It's aware of the navigation hazards from both ground and ceiling conditions considering the locomotion and height adjustment capabilities of the robot.
Both the tomogram construction and the scene evaluation process are accelerated through parallel computation.
Third, we reduce the burden of path planning by searching through multiple 2.5D tomogram slices, achieving higher efficiency than directly planning in 3D spaces.
Our main contributions include:
\begin{itemize}
    \item We propose a novel tomographic scene representation to understand 3D environments, extending elevation maps to multi-layer scenarios while maintaining their advantages in mapping efficiency and representation capabilities.
    \item We design a GPU-based tomographic map construction and scene evaluation method that reduces the computation time by 3 orders of magnitude while considering the robot's locomotion and height adjustment capabilities.
    \item We develop a trajectory generation approach that efficiently returns 3D trajectories with velocity information by searching and optimizing paths on tomogram slices, improving the path planning speed by 3 times.
    \item We integrate these modules to present a highly efficient navigation framework that improves the navigation speed by 2 orders of magnitude and successfully navigates a quadrupedal robot in real-world scenarios.
\end{itemize}

\section{Related Work}
In recent years, intensive research has been conducted on navigating ground robots in complex environments.
The majority of these approaches can be classified into three categories which respectively solve the navigation problems on point clouds, meshes, and 2.5D or 3D grid maps.

    \subsection{Navigating on Point Clouds}
    The 3D point clouds are extensively adopted in robotic applications to represent the environments \cite{10099031,9775940,10092189,10013761}.
    Some approaches directly solve navigation problems on point cloud maps.
    Liu \cite{roboticOnline} proposes a GPU-accelerated tensor voting framework to evaluate the environment's geometric features on the point cloud and calculate geodesic vectors on 3D structures.
    It then adopts k-NN to construct a navigation graph on the point cloud and uses Dijkstra for path planning. 
    Krüsi \textit{et al.} \cite{drivingPoint} analyze the point distributions in local patches for terrain assessment and directly generate optimized trajectories on point clouds, enabling navigation on irregular terrains in multi-level facilities.
    Waibel \textit{et al.} \cite{howRough} evaluate the terrain roughness in off-road scenarios by applying a Convolutional-LSTM network on point cloud patches to predict the IMU responses.
    Although these methods are free of terrain surface reconstruction, the unordered point cloud data increases the complexity of scene understanding and trajectory generation, which may lead to low navigation efficiency. 

    \subsection{Navigating on Meshes}
    Meshes are widely used in computer graphics to model 3D structures as polygon surfaces, which are also used in 3D navigation problems.
    Ruetz \textit{et al.} \cite{ovpcMesh} develop OVPC mesh which generates watertight 3D meshes from point clouds and classifies the traversable spaces by calculating the direction of surface normal vectors, enabling path planning under overhangs.
    Brandão \textit{et al.} \cite{gaitMesh} propose GaitMesh for navigating quadrupedal robots in large-scale multi-floor structures. 
    A distance field is built from voxelized polygons considering the properties of different gait controllers and a navigation mesh is then reconstructed for path searching using A* \cite{astar}.
    Pütz \textit{et al.} \cite{spvfMesh} plan paths on triangular meshes with arbitrary shapes in outdoor environments.
    It measures multiple geometric attributes of the mesh to generate the layered mesh map and adopts \ac{FMM} to compute the travel costs through wavefront propagation.
    However, these methods still usually require a long computation time to generate and evaluate the navigation mesh in large-scale scenarios.
    
    \subsection{Navigating on Grid Maps}
    Grid maps have high compatibility with a wide range of both traditional and learning-based approaches for mapping \cite{octomap,hoeller2022neural,stolzle2022recon,yang2023neural}, scene evaluation \cite{miki2022learning,yang2022GTICN,fu2022coupling,huang2023FAEL}, and path planning \cite{guzzi2020motion,lorenz2021rough,yang2021RTOptimal,cai2022prob}.
    For example, elevation maps \cite{elevation2014,elevation2018} are 2.5D grid maps that represent terrain structures with continuous height values and are successfully applied to solve navigation problems on complex terrains in \cite{guzzi2020motion,yang2021RTOptimal}. 
    Nevertheless, traditional elevation maps are unsuitable for scenarios with overhanging objects or multi-floor structures.
    Miki \textit{et al.} \cite{elevationCupy} use ramp parameters to exclude the ceiling points and update the local map with overlap clearance when moving through multiple floors.
    Meng \textit{et al.} \cite{terrainNet} introduce a ceiling layer and two ground elevations to capture the overhanging objects, ground obstacles, and terrain conditions, which enables traveling in complex off-road environments.
    However, these approaches are still unsuitable for global navigation in multi-layer environments. 
    Triebel \textit{et al.} \cite{MLSmap} record the height and thickness of various surface patches into the list of each grid to represent multi-layers scenarios, which may bring difficulties for the map processing stage to evaluate the scene from lists.

    Some approaches represent the environment using 3D grids.
    Frey \textit{et al.} \cite{voxelTrav} adopt a neural network for traversability estimation on occupancy voxels and plan paths in 3D structures with ceilings, while the voxels only provide a rough view of the terrain conditions.
    Wang \textit{et al.} \cite{beyond2D} extract traversable regions from the point cloud using a "Valid Ground Filter".
    It further obtains the travel costs by constructing \ac{ESDF} and performing local plane fitting.
    The trajectories are directly planned and optimized on dense voxels, which may increase the computational burden. 

    This paper proposes a navigation framework for ground robots in complex 3D environments using a tomographic scene representation based on 2.5D grid maps.
    Compared with existing extensions of elevation maps \cite{elevationCupy,terrainNet,MLSmap}, our approach adapts to large-scale multi-layer scenarios while maintaining representation simplicity.
    Rather than directly solving the navigation problems on 3D representations, our approach achieves higher efficiency in map construction, scene evaluation, and trajectory generation than \cite{roboticOnline,spvfMesh,beyond2D}.
    
\section{Methodology}
This section introduces the concept of point cloud tomography and the approach to constructing the tomogram slices.
Then it illustrates how we achieve efficient scene understanding and trajectory generation using this new representation.

    \subsection{Tomogram Construction}
\begin{figure}[t]
\setlength{\abovecaptionskip}{0pt}
\centering
\includegraphics[width=3.0in]{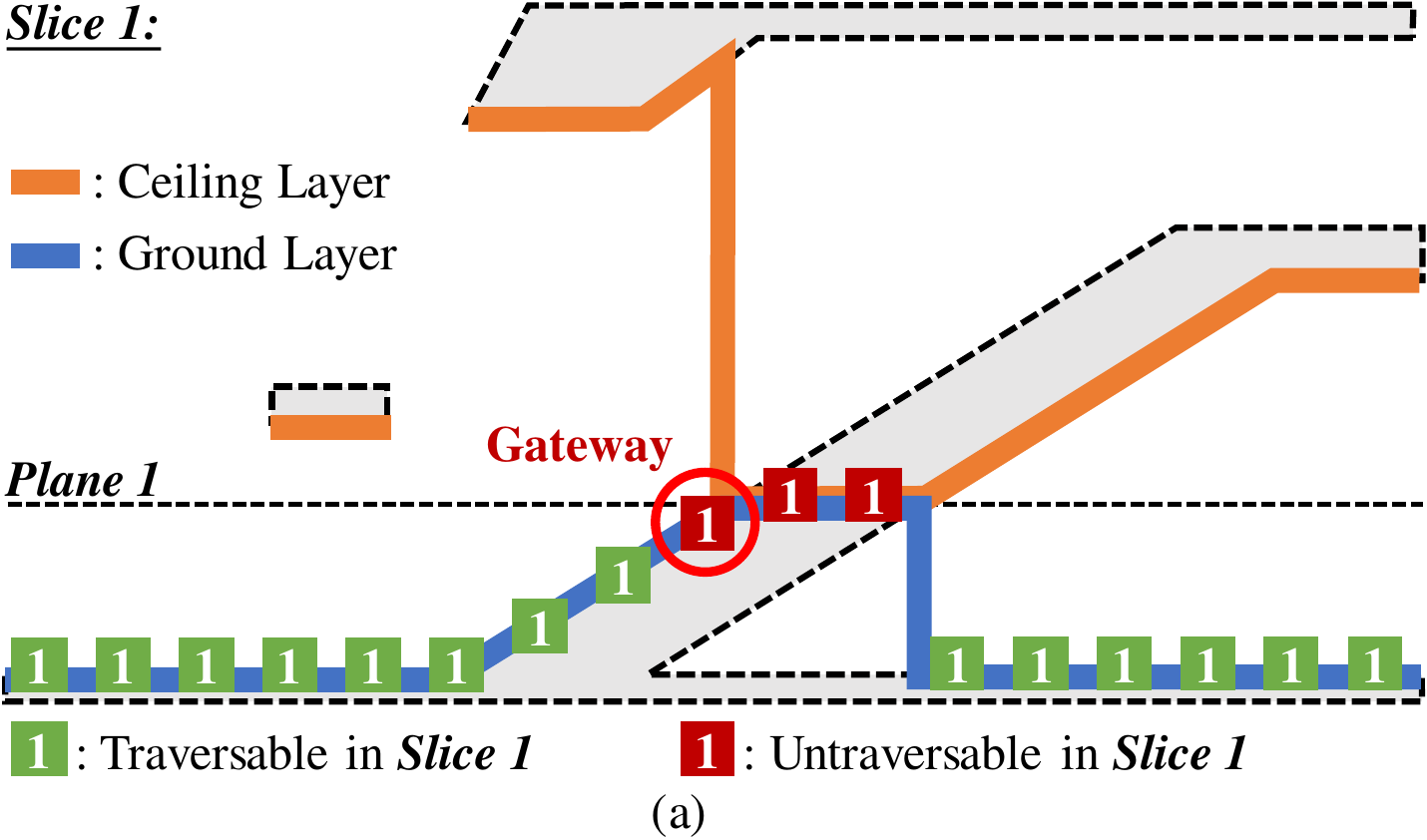} \vspace{0.5em}

\includegraphics[width=3.0in]{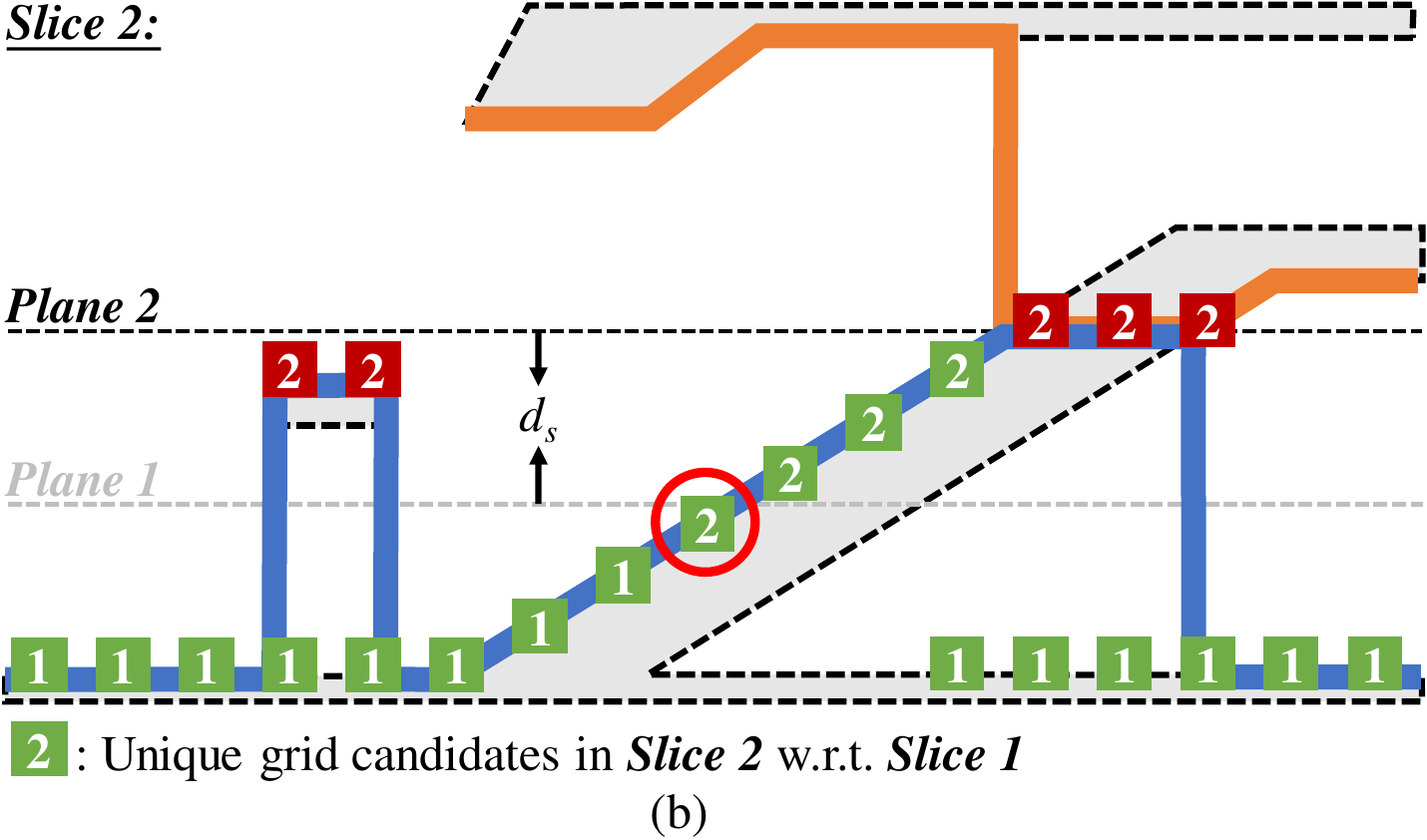} \vspace{0.5em}

\includegraphics[width=3.0in]{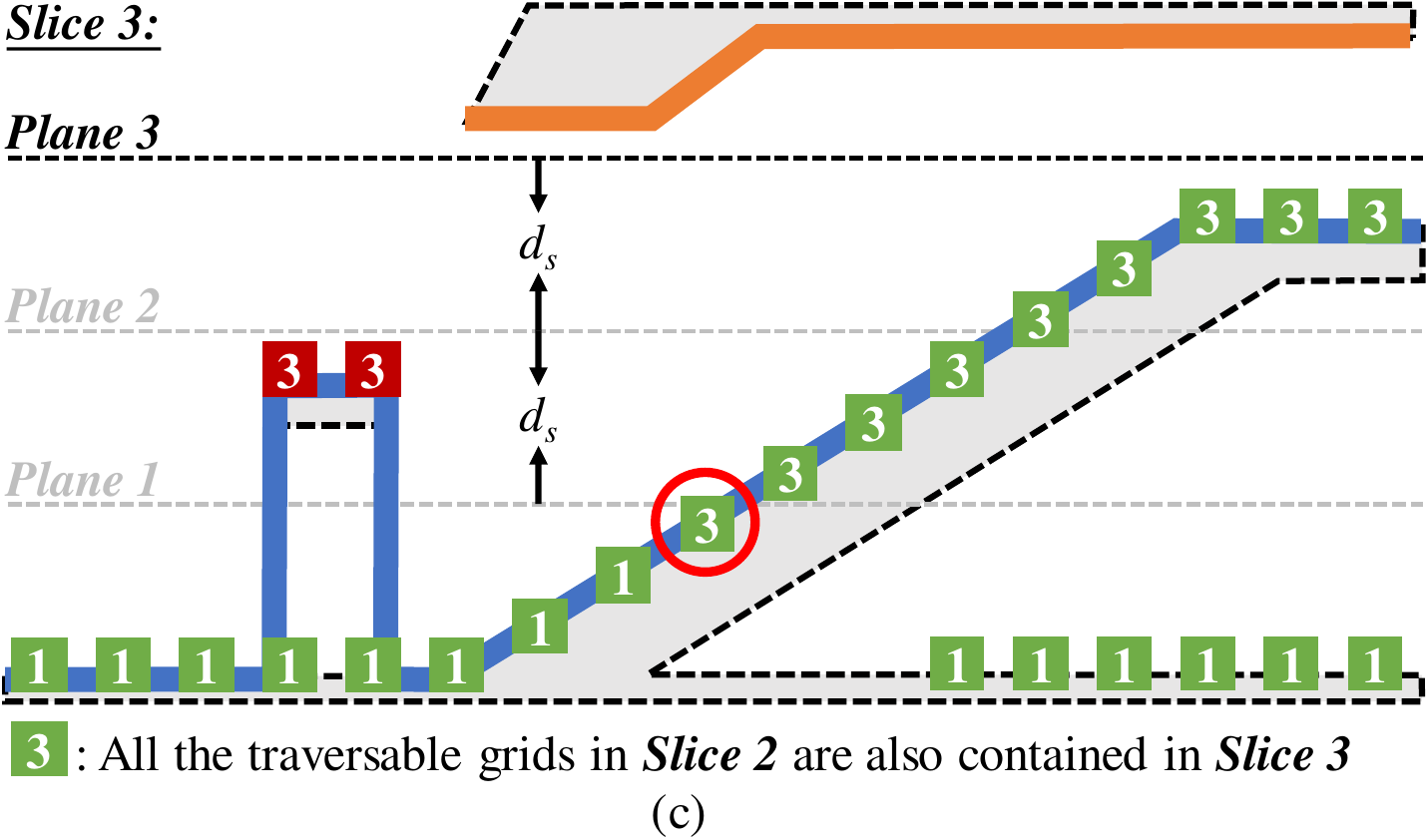}
\caption{
We construct tomogram slices by projecting the point cloud onto a series of equidistant horizontal planes. 
Each slice contains a ceiling (orange) and a ground (blue) layer.
The squares present the grid traversability (green: traversable, red: untraversable) considering the ground conditions and the ceiling height.
As all the traversable grids in slice 2 are contained in the union of slices 1 and 3, slice 2 can be omitted as done in Section \ref{sec:simp}.
The red-circled grids are the ``gateways" defined in Section \ref{sec:planning} that connect slice 1 to the upper slices for searching upwards, as these grids share the same spatial position but the cost in the upper slice is lower and reflects the real traversability at that position.
By using the gateways, our planner travels through multiple slices to search on the slope and under the overhangs, thus enabling navigation in multi-layer structures.
}
\label{fig:tomo}
\vspace{-1.0em}
\end{figure}

    We define the ``tomogram slices" $\{S_k \, | \, k \in [0, N]\}$ which are multi-channel grid maps with the resolution being $r_g$. 
    Each slice $S_k=\{\bm{e}_k^G, \bm{e}_k^C, \bm{c}_k^T\}$ contains a ground $\bm{e}_k^G=\{e_{i,j,k}^G\}$ and a ceiling $\bm{e}_k^C=\{e_{i,j,k}^C\}$ elevation layer that respectively encodes the terrain conditions and the overhanging structures, as well as a corresponding travel cost map $\bm{c}_k^T=\{c_{i,j,k}^T\}$, where $e,c$ are the elevation and cost values of each grid, $i,j$ are the row and column indices, $k$ is the slice index.
    To construct the tomogram, we analyze the point cloud from a series of cross-sections viewed from the relevant horizontal planes (Fig. \ref{fig:tomo}). 
    The planes are equidistantly stacked with the separation being $d_s$. 
    The first plane is placed above the lowest point in the point cloud by $d_s$ and the number of planes $N$ is selected to ensure that the last plane is above the highest point. 

    Each plane splits the point cloud into the ``lower" and the ``upper" group containing the points below and above the plane.
    As shown in Algorithm \ref{alg:pct} (line 1 to 6), to construct the ground layer $\bm{e}_k^G$ of the $k$-th slice, we vertically project each point in the ``lower group" upwards to the plane and obtain its projection depth by calculating the height difference between the point and the plane.
    The projected points are rasterized to obtain a grid map.
    For each grid, the ground elevation $e_{i,j,k}^G$ is the plane height minus the minimum projection depth among all the points inside.
    The ceiling $\bm{e}_k^C$ is obtained similarly by projecting the ``upper group" downwards and $e_{i,j,k}^C$ is the plane height plus the minimum projection depth.
    We set $e_{i,j,k}^G$ or $e_{i,j,k}^C$ to be invalid if no point is projected into the grid.
    In this way, these planes generate pairs of ground and ceiling layers for the respective tomogram slices, while the travel cost maps $\bm{c}_k^T$ are further obtained in Section \ref{sec:trav_estm}.
    
    We note that in an indoor scenario, the building floors and our slices are not necessarily aligned, as a single floor may also contain various height layers.
    In addition, our approach applies to more complicated structures such as spiral overpasses or caves with intricate passages.
    To ensure complete coverage of the valid planning search space, we choose $d_s \leq d_{min}$, where $d_{min}$ is the minimum ground-ceiling interval for the robot to move through. 
    Therefore, all the places with sufficiently large intervals would be split by at least one plane.
    Then the traversable places will be recognized by evaluating the relevant ground and ceiling elevations in Section \ref{sec:trav_estm}.

\begin{algorithm}[t]
    \caption{Point Cloud Tomography}
    \label{alg:pct}
    \textbf{Input:} global point cloud map $P=\{\bm{p}_u|\bm{p}_u=[x_u, y_u, z_u]^T\}$ \\
    \textbf{Output:} tomogram slices $S=\{S_k|S_k=(\bm{e}_k^G, \bm{e}_k^C, \bm{c}_k^T)\}$
    \begin{algorithmic}[1]
        \State Minimum height of points $z_{min}=\min{\{z_u\}}$
		\For {each point $\bm{p}_u=[x_u, y_u, z_u]^T$}
            \State Grid index $i,j=$ \textbf{rasterize}$(x_u, y_u)$
            \For {slice index $k=0,1,\ldots,N$}
                \State $e_{i,j,k}^G=\max{(z_u, e_{i,j,k}^G)}$ \textbf{if} $z_u \geq z_{min} + k d_s$
                \State $e_{i,j,k}^C=\min{(z_u, e_{i,j,k}^C)}$ \textbf{if} $z_u < z_{min} + k d_s$
            \EndFor
		\EndFor
        \State $\bm{c}_k^{init}=\textbf{travEstm}(\bm{e}_k^G, \bm{e}_k^C)$ (Eq. \ref{eq:trav_estm_1} to \ref{eq:trav_estm_5})
        \State $\bm{c}_k^T=\textbf{inflation}(\bm{c}_k^{init})$ (Eq. \ref{eq:inflation})
        \For {each slice $S_k=(\bm{e}_k^G, \bm{e}_k^C, \bm{c}_k^T)$ in $S=\{S_k\}$}
            \State Find unique grids $U_k=\{\textbf{unique}(\bm{e}_k^G,\bm{c}_k^T)\}$ (Eq. \ref{eq:simp_l}, \ref{eq:simp_u})
            \State \textbf{if} $\textbf{size}(U_k)=0$, \textbf{remove} $S_k$ from $S$
        \EndFor
		\State \Return {unique tomogram slices $S$}
	\end{algorithmic} 
\end{algorithm}

    \subsection{Traversability Estimation} \label{sec:trav_estm}
    Next, we estimate the scene traversability and calculate the grid-wise travel costs by analyzing the ground and ceiling layers in the constructed tomogram (Algorithm \ref{alg:pct} line 7, 8).
    Compared with existing scene evaluation methods using single terrain elevations, our approach is also aware of the overhanging hazards and supports active body height adjustment of the robot.
    To achieve this, we calculate the height difference between each pair of ground and ceiling layers to get the interval distance $\bm{d}^I=\bm{e}^C-\bm{e}^G$.
    Suppose the robot's body height is adjustable between $d_{min}$ and a normal operation height $d_{ref}$, a grid is considered untraversable if $d^I < d_{min}$ or will be assigned a penalty for the additional height adjustment effort if $d^I \in [d_{min}, d_{ref}]$.
    For instance, we use the following interval cost map $\bm{c}^I$ for our quadrupedal robot as lowering the body height leads to more energy consumption while walking, where $c^B$ is the cost for untraversable barriers, $\alpha_d$ is a scaling factor, and each element $c^I$ is obtained as follows:
    \begin{gather} \label{eq:trav_estm_1}
        c^I = 
        \begin{cases}
            c^B & \text{ if } d^I < d_{min} \\
            \max\left(0, \ \alpha_d (d_{ref}-d^I)\right) &\text{ otherwise}
        \end{cases}.
    \end{gather}
    By integrating $\bm{c}^I$ into the cost map, our planner generates cost-optimal 3D trajectories in Section \ref{sec:planning} and \ref{sec:optimization} considering the overhangs even though the robot's motion for height adjustment is decomposed from planning on the ground surfaces.

    Next, we analyze the ground layers to obtain the cost maps of terrain conditions $\bm{c}^G$.
    Compared with \cite{beyond2D}, our approach also identifies traversable steps and stairs to navigate legged robots.
    For each grid, we obtain the gradients $[g^x, g^y]^T$ of the ground elevation $e^G$ in $x, y$ directions through the finite-difference method and get the following magnitudes:
    \begin{gather} \label{eq:trav_estm_2}
        m^{xy} = \max(|g^x|, \ |g^y|), \ m^{grad} = \sqrt{(g^x)^2 + (g^y)^2},
    \end{gather}
    and we have $m^{xy} \leq m^{grad}$.
    Three criteria with thresholds $\theta_b, \theta_s, \theta_p$ are applied to measure the terrain conditions. 
    The grid is considered the boundary of a barrier if $m^{xy} > \theta_b$ or a traversable gentle surface if $m^{grad} < \theta_s$, and the cost element:
    \begin{gather} \label{eq:trav_estm_3}
        c^G = 
        \begin{cases}
            c^B & \text{ if } m^{xy} > \theta_b \\
            \alpha_s \left(\frac{m^{grad}}{\theta_s}\right)^2 & \text{ if } m^{grad} < \theta_s
        \end{cases},
    \end{gather}
    where $\alpha_s$ is a scaling factor.
    Otherwise, the grid belongs to an edge that might be untraversable for wheeled robots (set $c^G=c^B$) but can still be stepped across by legged robots.
    We further estimate if the surrounding grids are safe to step on by calculating the percentage $p_s$ of neighboring grids with $m^{grad} < \theta_s$ within a local patch.
    The grid is considered traversable if $p_s > \theta_p$ and the cost is formulated as:
    \begin{gather} \label{eq:trav_estm_4}
        c^G = 
        \begin{cases}
            \alpha_b \left(\frac{m^{xy}}{\theta_b}\right)^2 & \text{ if } p_s > \theta_p \\
            c^B & \text{ otherwise}
        \end{cases}.
    \end{gather}
    
\begin{figure}[t]
\setlength{\abovecaptionskip}{0pt}
\centering
\includegraphics[width=3.45in]{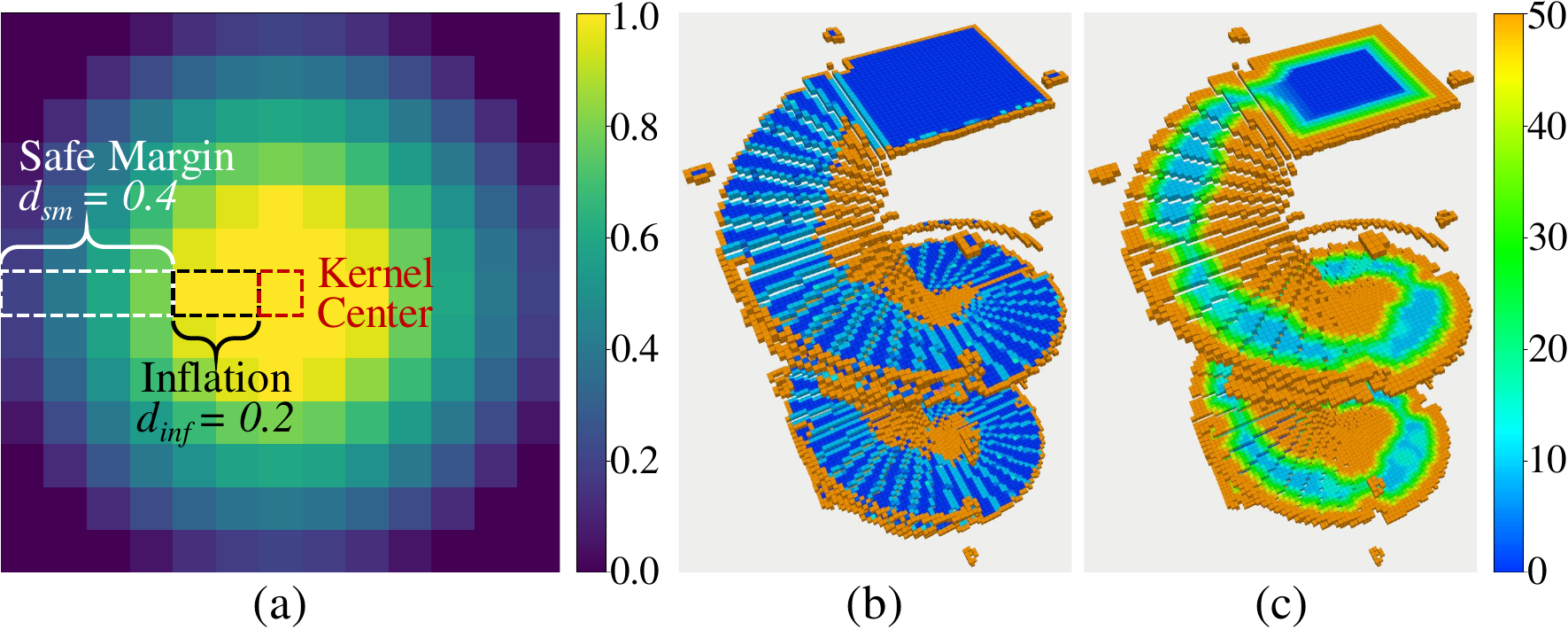}
\caption{
(a): Illustration of the inflation kernel when $d_{inf}=0.2$, $d_{sm}=0.4$ and $r_g=0.1$. 
(b): The traversability cost map before inflation, where the untraversable grids are in orange and the traversable regions are in blue. 
The regions near the spiral center are untraversable due to the insufficient stair width.
(c): The final travel cost map, where the untraversable regions are inflated and the costs are gradually reduced within the safe margin after applying the inflation kernel. 
}
\label{fig:kern_inf}
\vspace{-1.0em}
\end{figure}

    As shown in Fig. \ref{fig:kern_inf} (b), the grids near the center of the spiral stairs are untraversable as the surfaces are too narrow. 
    The initial cost map is the clipped sum of $\bm{c}^I$ and $\bm{c}^G$:
    \begin{gather}  \label{eq:trav_estm_5}
        \bm{c}^{init} = \min(c^B, \ \bm{c}^I + \bm{c}^G).
    \end{gather}

\begin{figure*}[t]
\setlength{\abovecaptionskip}{0pt}
\centering
\includegraphics[width=6.5in]{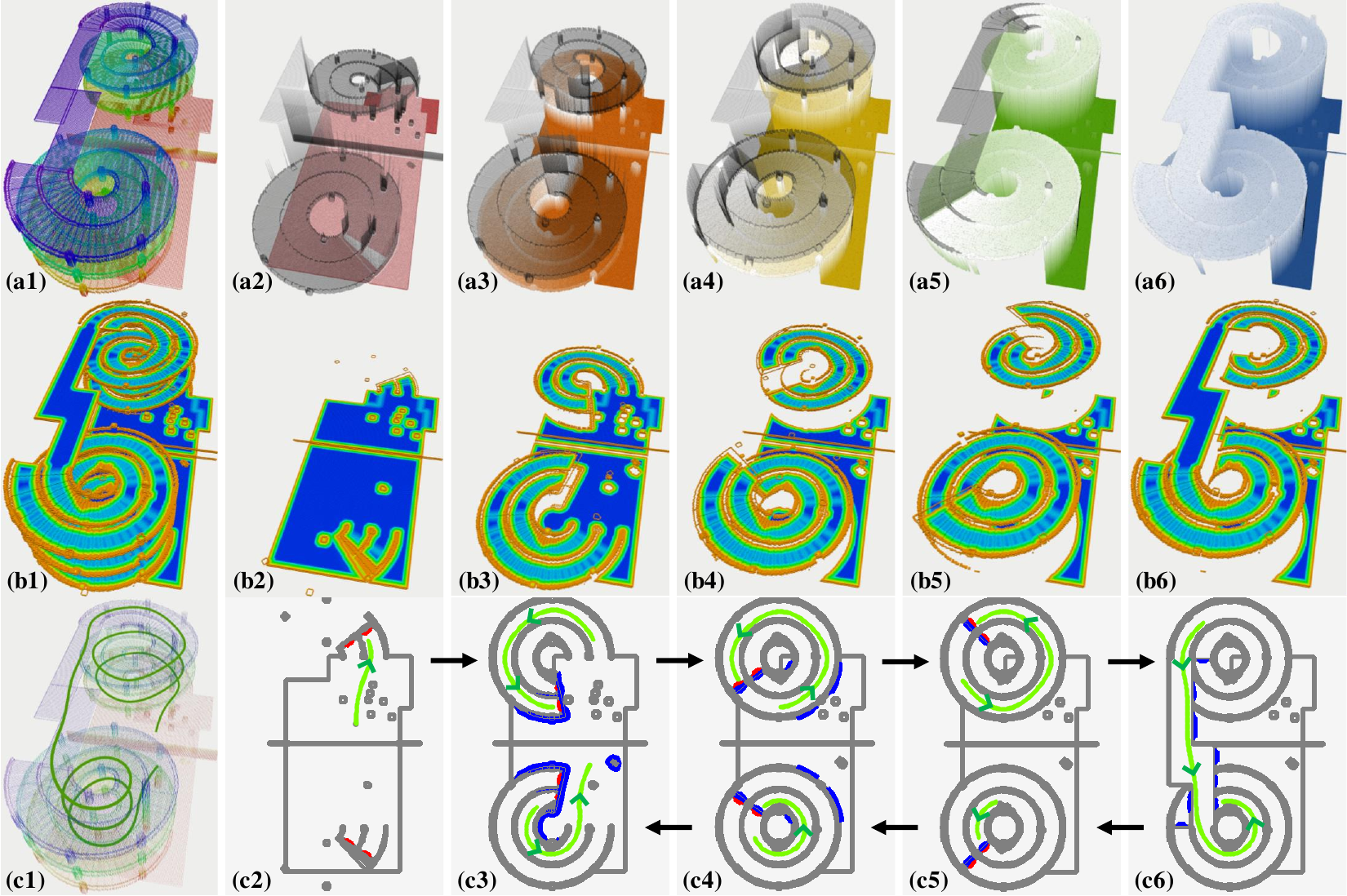}
\caption{
(row a): The point cloud map (a1) of the \textit{spiral} environment in \cite{beyond2D} is used to construct our tomogram.
The environment is then represented by five 2.5D tomogram slices (a2) to (a6) after simplification, where the gray layers are ceilings and the colored layers are ground elevations. 
(row b): The whole planning space (visualized as a point cloud (b1), blue: traversable, orange: untraversable) can be obtained by integrating (b2) to (b6) which are 2.5D travel cost maps with the related ground elevations in (a2) to (a6).
Note that (b1) is only for visualization and we directly adopt (b2) to (b6) for trajectory generation in Section \ref{sec:planning} and \ref{sec:optimization}.
(row c): The planner searches on multiple tomogram slices along the green arrows and generates the 3D trajectory. It can enter the upper slices through the gateways in red and can also search downwards through the gateways in blue.
}
\label{fig:spiral}
\vspace{-1.0em}
\end{figure*}   

    Finally, we apply an inflation kernel (Fig. \ref{fig:kern_inf} (a)) on $\bm{c}^{init}$ to expand the untraversable regions by $d_{inf} \geq r_c$ considering the robot's collision radius $r_c$ and create a safe margin within $d_{sm}$ for smooth cost gradients and safe navigation behaviors.
    The weight $K(m,n)$ of a kernel cell at $(m,n)$ with resolution $r_g$ is calculated based on the Euclidean distance $d_{mn}$ between the kernel center and the cell:
    \begin{gather} \label{eq:inflation}
        K(m,n)=\max \left(0, \ \min\left(1-\frac{d_{mn}-d_{inf}}{d_{sm}-r_g}, \ 1\right)\right).
    \end{gather}
    
    The final travel cost map $\bm{c}_k^T$ is obtained through the sliding window method, where the kernel performs the Hadamard product with the patches on $\bm{c}^{init}_k$ and returns the maximum value of each resulting matrix as the cost of the patch center.
    
    \subsection{Tomogram Simplification} \label{sec:simp}
    During the tomogram construction, the planes are separated by $d_s=d_{min}$ to guarantee complete coverage of the planning space.
    However, such a small height increment brings quite a limited expansion of the mapping region in the new layer, resulting in low efficiency in path searching or map storage. 
    Therefore, the tomogram is further simplified (Algorithm \ref{alg:pct} line 9 to 11) based on the following principle:
    Let $M_k$ denote the set containing all the traversable grids in slice $S_k$. If:
    \begin{gather} \label{eq:tomo_simp}
        M_k \subset (M_{k-1} \cup M_{k+1}),
    \end{gather}
    which means the search space of an intermediate slice $S_k$ is already included in the union of its previous and subsequent slice, then $S_k$ is redundant and can be omitted.
    If so, $S_{k+1}$ will become the new intermediate slice and we then check if $M_{k+1} \subset (M_{k-1} \cup M_{k+2})$.
    Otherwise, $S_k$ will be preserved and we continue to examine if $M_{k+1} \subset (M_{k} \cup M_{k+2})$.
    The process is repeated until all the slices are checked.

    The condition in Eq. \ref{eq:tomo_simp} can be checked according to the ground elevations and the travel costs.
    A traversable grid with $c_{i,j,k}^T < c^B$ at at position $(i,j,k)$ is considered ``unique" if:
    \begin{gather}
        \label{eq:simp_l} (e_{i,j,k}^G - e_{i,j,k-1}^G > 0 \text{ or } c_{i,j,k}^T < c_{i,j,k-1}^T) \text{ and} \\
        \label{eq:simp_u} (e_{i,j,k+1}^G - e_{i,j,k}^G > 0 \text{ or } c_{i,j,k}^T < c_{i,j,k+1}^T),
    \end{gather}
    where Eq. \ref{eq:simp_l} indicates that the grid has a unique spatial position or has the same position but a lower cost that reflects its real traversability compared with the grid at $(i,j,k-1)$.
    Similarly, Eq. \ref{eq:simp_u} checks the relationship with the grid in $S_{k+1}$.
    In this way, we examine the uniqueness of all the grids in each slice and the slices containing no unique grid will be omitted.
    For instance, in Fig. \ref{fig:tomo} (b) the traversable grids of slice 2 on the slope (green grids with a mark ``2") satisfy Eq. \ref{eq:simp_l} and are considered as candidates of unique grids w.r.t. slice 1. 
    However, they violate Eq. \ref{eq:simp_u} as they have the same elevations and costs as the grids in slice 3. 
    Therefore, slice 2 can be omitted.
    
    Fig. \ref{fig:spiral} shows the simplified tomogram (row a) of the \textit{spiral} scenario in \cite{beyond2D} and the corresponding travel cost maps (row b). 
    The point cloud is initially projected onto 46 slices which are then simplified to 5 slices (Fig. \ref{fig:spiral} (a2) to (a6)) using the above procedures.
    Compared with storing travel costs into dense 3D voxels as done in \cite{beyond2D}, our approach achieves higher efficiency by only adopting 5 layers of 2.5D cost maps (Fig. \ref{fig:spiral} (b2) to (b6)) to represent the whole planning space.
    The memory usage of our tomogram has the potential to be further reduced by adopting sparse representations or other data structures to store the unique grids only.
    In this paper, we continue to use multi-layer 2.5D grid maps to represent the environment for the simplicity of the subsequent processing stages.
    
    \subsection{Path Planning through Slices} \label{sec:planning}
    After obtaining the simplified travel cost maps, we modify A* to search through tomogram slices and plan initial paths.
    Rather than directly planning on dense voxels in \cite{beyond2D}, we separately plans on the 2.5D maps and optimizes $z$-axis motions to avoid overhangs, achieving higher efficiency.
    As the cost map already contains the interval cost $\bm{c}^I$, our planner generates optimal solutions considering the ground-ceiling intervals.

\begin{figure*}[t]
\setlength{\abovecaptionskip}{0pt}
\centering
\includegraphics[width=6.5in]{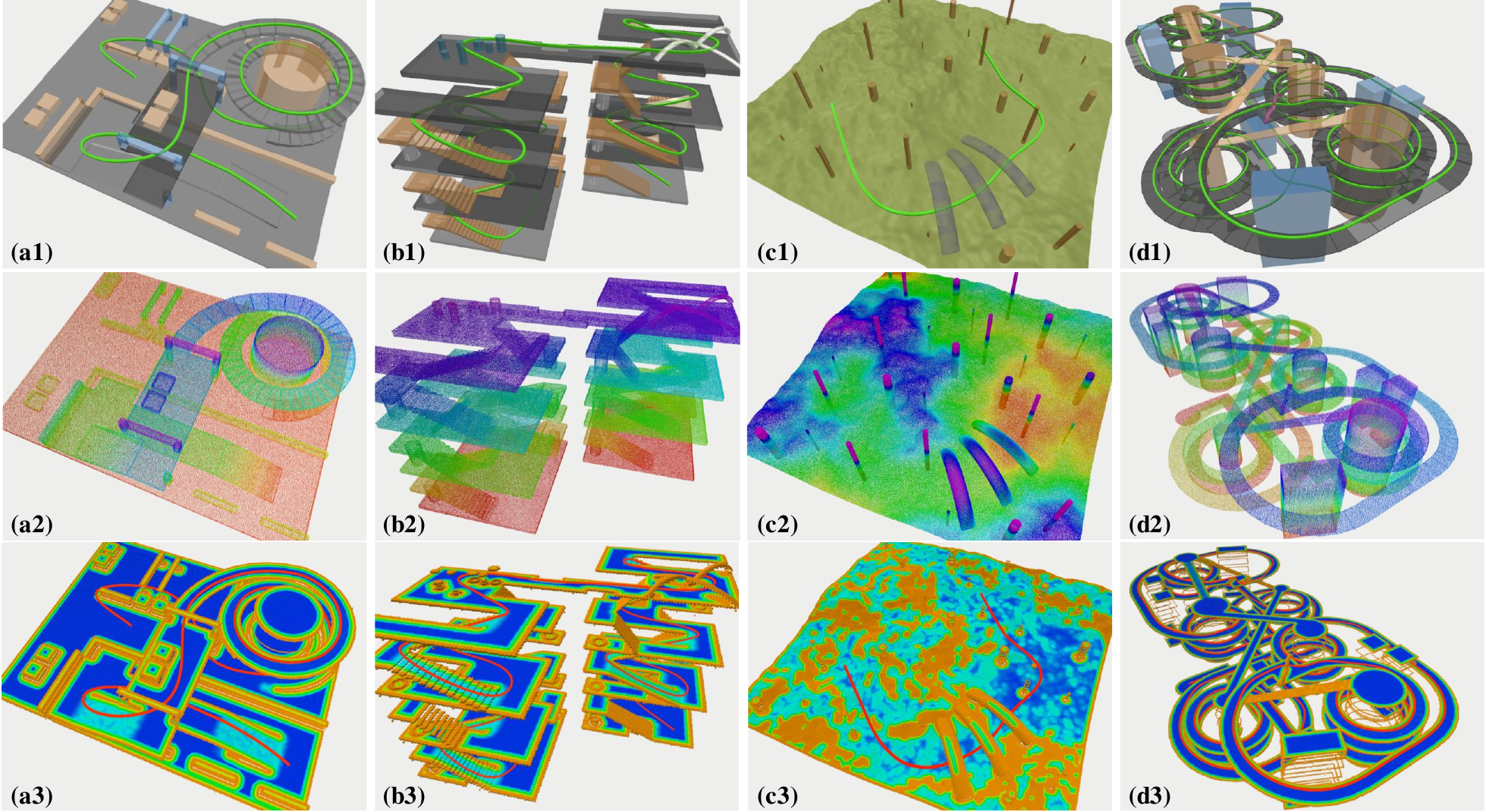}
\caption{
The simulation scenarios and the trajectories generated by our approach. 
The first row presents the 3D models of environments: \textit{factory} (a1), \textit{building} (b1), \textit{forest} (c1), and \textit{overpass} (d1). 
The second row shows the point clouds of the corresponding scenarios and the third row presents the integrated travel cost maps generated by our approach.
Our approach provides feasible and smooth trajectories in these complex 3D scenarios with various structured or irregular terrain features and overhanging objects.
}
\label{fig:tomo_sim}
\vspace{-1.0em}
\end{figure*}

    Given an initial and a goal position, the planner starts searching on an arbitrary slice $S_k$ that contains the initial position.
    Each grid is considered a graph node that connects to its eight neighbors on the same cost map.
    When a grid is queried, we also check the two grids from the adjacent slices $S_{k-1}, S_{k+1}$ at the same planimetric position $(i,j)$.
    Taking the grid above at $(i,j,k+1)$ as an example, if they share the same ground elevation $e_{i,j,k}^G = e_{i,j,k+1}^G$, then they are considered the same node in the 3D space with the node cost $c^N = \min(c_{i,j,k}^T, \, c_{i,j,k+1}^T)$. 
    The cost for graph search between two nodes is defined as $c^N$ of the target node plus their Euclidean distance, where the connection breaks if $c^N=c^B$.
    The heuristic cost is the diagonal distance from the queried node to the goal.
    If $c_{i,j,k+1}^T < c_{i,j,k}^T$, then this node is considered as a ``gateway" (e.g., red-circled grids in Fig. \ref{fig:tomo}) to travel from $S_k$ to $S_{k+1}$ so that the planner can continue to search on $S_{k+1}$.
    The same logic adapts to the grid below at $(i,j,k-1)$ when searching downwards, where the gateway occurs if $e_{i,j,k-1}^G = e_{i,j,k}^G$ and $c_{i,j,k-1}^T < c_{i,j,k}^T$.
    In this way, our approach plan paths on different slices and connect the paths at the gateways to generate a 3D result on the multi-layer structure.
    Fig. \ref{fig:spiral} (row c) presents how our planner travels on multiple slices along the green arrows. 
    The planner travels to the upper slices through the gateway grids in red and enters the lower slices through the gateway grids in blue.

    \subsection{Trajectory Optimization} \label{sec:optimization}
    The proposed scene representation is compatible with various gradient-based trajectory optimization techniques. 
    In this paper, we represent the trajectory as $M$-piece 3-dimensional polynomials. 
    The $i$-th piece of the trajectory is defined by a polynomial with a degree of $N=5$: 
    \begin{equation}
        \bm{q}_i(t) = \bm{\sigma}_i^T \bm{\beta}(t),\, \forall t \in \left[0, T_i\right],
    \end{equation}
    where $\bm{\sigma}\in \mathbb{R}^{(N+1)\times m}$ is the coefficient matrix and $\bm{\beta}=\left[1, t, \dots, t^N\right]^T$ is the natural bias. 
    The trajectory optimization is formulated as a minimal control effort problem:
    \begin{subequations}
    \begin{align}
        \underset{\bm{\sigma}, T}{\min}~&~J_c + w_z ||q_z(t) - Z_{ref}\left(\bm{q}(t)\right)||_2 + w_T T \\
        \mathit{s.t.}~&~\bm{q}_1(0) = \bar{\bm{q}}_0,~\bm{q}_{M}(T) = \bar{\bm{q}}_f,\\
        &~\bm{q}^{[3]}_{i}(T_i) = \bm{q}^{[3]}_{i+1}(0),~i = \{1,\dots, M-1\}, \\
        &~\mathcal{C}(\bm{q}(t)) \geq C_{safe},~\forall t \in [0, T],\\
        &~\mathcal{G}(q(t),\dots,q^{(2)}(t)) \preceq \bm{0},~\forall t \in [0, T], \\
        &~ \mathcal{H}_g\left(\bm{q}(t)\right) \leq q_{z}(t) \leq \mathcal{H}_c\left(\bm{q}(t)\right),~ \forall t \in [0, T],
    \end{align}
    \end{subequations}
    where $J_c$ is the jerk control effort to smooth the trajectory in the 3D space following \cite{wang2022geometrically, cheng2022real}, $Z_{ref}\left(\bm{q}(t)\right)=e^G\left(\bm{q}(t)\right)+d_{ref}$ is the preferred operation height of the robot, $\mathcal{C}$ is the travel cost generated in Section \ref{sec:trav_estm}, $C_{safe}$ is the safety margin, and $\mathcal{G}(\cdot)$ are kinematic constraints on the maximum value of velocity, acceleration, and the changing rate of headings. 
    For ground robots with adjustable heights, additional maximum height constraints are imposed on the $z$-axis of the trajectory, where $\mathcal{H}_g\left(\bm{q}(t)\right)=e^G\left(\bm{q}(t)\right)+d_{min}$ and $\mathcal{H}_c\left(\bm{q}(t)\right)$ is the inflated ceiling elevation queried from the tomogram. 

\section{Experiments}
    \subsection{Implementation Details} \label{sec:imple}
        \subsubsection{Evaluation scenarios}
\begin{figure}[t]
\setlength{\abovecaptionskip}{0pt}
\centering
\includegraphics[width=3.4in]{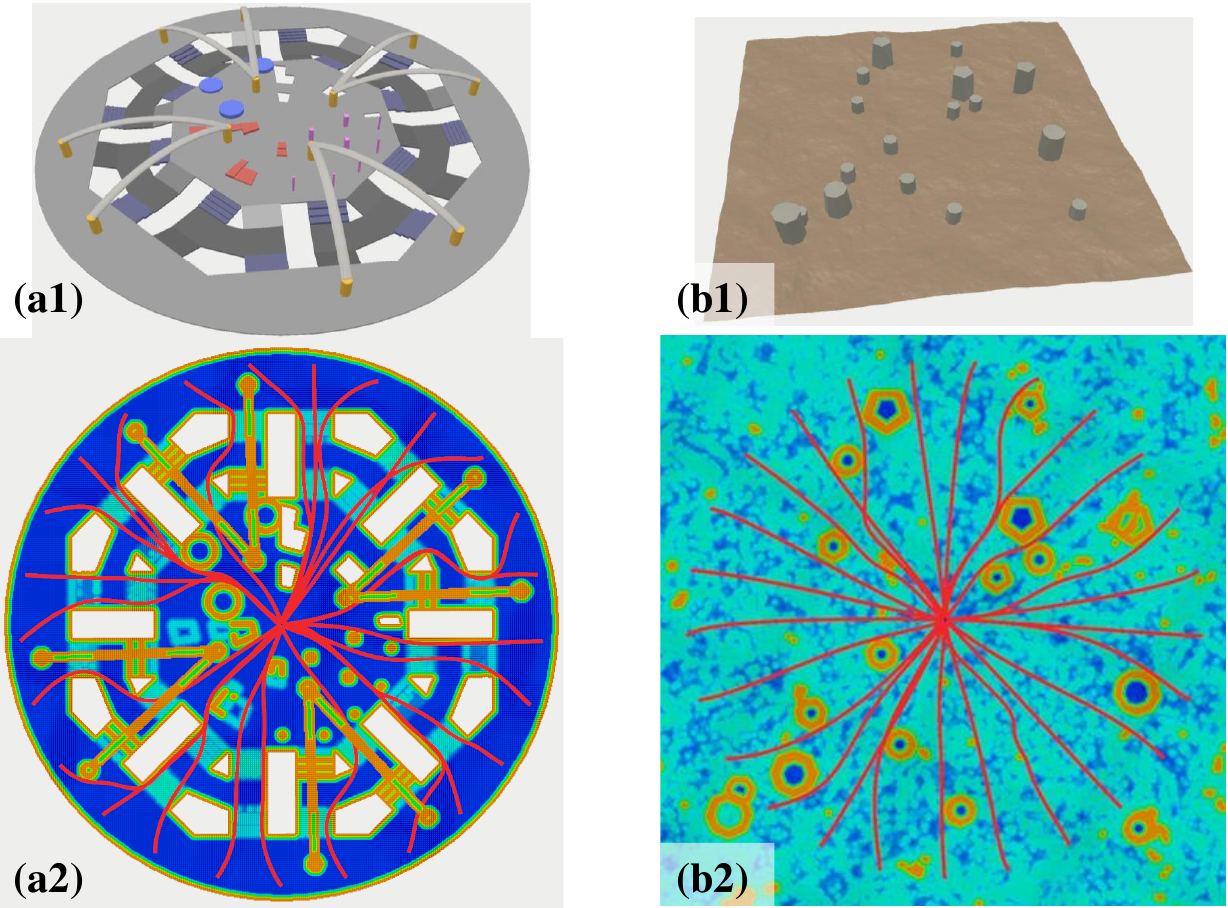}
\caption{
The \textit{plaza} (a1) scenario with multiple urban structures and the \textit{hills} (b1) scenario with irregular terrain and obstacles for the evaluation of trajectory quality. 
(b1), (b2): example trajectories generated by our approach.
}
\label{fig:traj50}
\vspace{-1.2em}
\end{figure}

        The navigation approaches are evaluated in simulated and real-world scenarios with complex 3D structures.
        In simulation, the following environments are prepared to analyze the navigation performance, where the dimensions are presented in (\textit{length}, \textit{width}, \textit{height}):
        \begin{itemize}
            \item \textit{Factory}: An outdoor scenario with regular urban terrain features and overhanging structures (Fig. \ref{fig:tomo_sim} (a1)).
            The scene dimension is $(84 \times 68 \times 12)\unit{\m}$.
            \item \textit{Building}: An indoor scenario containing multiple floors with stairs and slopes of different steepness (Fig. \ref{fig:tomo_sim} (b1)). 
            The scene dimension is $(22 \times 20 \times 16)\unit{\m}$.
            \item \textit{Forest}: A wilderness environment with a tunnel, irregular terrain, and thin obstacles (Fig. \ref{fig:tomo_sim} (c1)).
            The scene dimension is $(40 \times 40 \times 7)\unit{\m}$.
            \item \textit{Overpass}: A large-scale spiral overpass with multiple layers and a complex route (Fig. \ref{fig:tomo_sim} (d1)).
            The scene dimension is $(155 \times 95 \times 30)\unit{\m}$.
        \end{itemize}
        The resolution of scene representations is $0.2$ in \textit{factory, overpass} and $0.1$ in \textit{building, forest} to capture the terrain features.
        To validate the trajectories, we simulate a Pioneer 3-DX robot in CoppeliaSim \cite{coppeliaSim} for \textit{factory}, \textit{forest}, and \textit{overpass}.
        In addition, to evaluate the trajectory generation performance, we uniformly sample $50$ goals that are $26.5\unit{\m}$ away from the starting point at the map center in each of the following scenarios:
        \begin{itemize}
            \item \textit{Plaza}: An outdoor scenario with various urban structures (Fig. \ref{fig:traj50} (a1)).
            The scene dimension is $(56 \times 56 \times 5)\unit{\m}$.
            \item \textit{Hills}: A wilderness scenario with irregular obstacles and rough terrain (Fig. \ref{fig:traj50} (b1)). 
            Dimension: $(60 \times 60 \times 3)\unit{\m}$.
        \end{itemize}

        In real-world experiments, we scan the following environments with LiDARs and build the point cloud maps using A-LOAM\footnote{https://github.com/HKUST-Aerial-Robotics/A-LOAM} \cite{zhang2014loam}, where we manually remove the noise points and dynamic objects for rational navigation behaviors:
        \begin{itemize}
            \item \textit{Stairs}: A multi-layer scenario where the robot starts from the ground floor and moves up the winding stairs (Fig. \ref{fig:tomo_real} (a1)). The scene is scanned by a Livox Mid-70.
            The scene dimension is $(14 \times 14 \times 7)\unit{\m}$.
            \item \textit{Boxes}: A sandbox where the robot moves through narrow arches and reaches the upper platform from the stairs or the slope (Fig. \ref{fig:tomo_real} (b1)). The scene is scanned by an RS-Helios-32 onboard the robot.
            Dimension: $(22 \times 22 \times 4)\unit{\m}$.
        \end{itemize}
        We use a Jueying Mini \cite{Jueying} quadrupedal robot to validate the trajectories generated by the proposed framework.
        The algorithm parameters are set considering the motion capabilities of robots.
        For the quadrupedal robot used in the experiments, the parameter values in Table \ref{tab:params} are adopted for map construction and traversability estimation.
        For wheeled robots, we can set $\theta_p=1.0$ to disable planning over stairs or steps.

\renewcommand{\arraystretch}{1.0}
\renewcommand\tabcolsep{5.5pt}
\vspace{-0.5em}
\begin{table}[h]
\caption{Algorithm Parameters}
\vspace{-0.5em}
\centering
\begin{threeparttable}
\begin{tabular}{l|cc}
    \toprule
    Items                     & Notations                              & Values \\ 
    \midrule 
    Height Adaptation         & [$d_s, \ d_{min}, \ d_{ref}$]               & [$0.50, \ 0.50, \ 0.65$]         \\
    Thresholds                & [$\theta_b, \ \theta_s, \ \theta_p$]        & [$1.70, \ 0.36, \ 0.20$]         \\
    Costs and Scaling Factors & [$c^B, \ \alpha_d, \ \alpha_b, \ \alpha_s$] & [$50, \ 20, \ 20, \ 15$] \\
    \bottomrule
\end{tabular}
\footnotesize
\end{threeparttable}
\label{tab:params}
\vspace{-0.5em}
\end{table}
\renewcommand{\arraystretch}{1.0}

        \subsubsection{Baseline approaches}
        Four methods with the following typical scene representations are implemented to compare the navigation capabilities with our approach.
        The last three methods which plan in multi-layer 3D structures are further quantitatively evaluated to compare the navigation efficiency in mapping, scene evaluation, and trajectory generation:
        \begin{itemize}
            \item \textit{Elevation maps}: Yang's method \cite{yang2021RTOptimal} is adopted to evaluate traditional elevation maps \cite{elevationCupy} on GPU and navigate quadrupedal robots on complex terrains using A*.
            \item \textit{Points}: Liu's approach \cite{roboticOnline} is used to perform GPU-accelerated tensor voting directly on point clouds for scene evaluation. The path search algorithm in the original implementation is replaced by A* for better efficiency. 
            \item \textit{Meshes}: Pütz's method \cite{spvfMesh} is adopted to evaluate the attributes of meshes and generate the layered navigation mesh map. The meshes are built from point clouds using the Ball-Pivoting algorithm. Then we also use A* to find paths on the mesh maps for comparison.
            \item \textit{Voxels}: We use Wang's method\footnote{https://github.com/ZJU-FAST-Lab/3D2M-planner} \cite{beyond2D} which processes the point clouds into \ac{ESDF} maps. The initial paths are obtained by searching on 3D grids using A* and then optimized to generate smooth trajectories.
        \end{itemize}
        
        We evaluate the navigation efficiency by comparing the time $T_p$ for pre-processing the point cloud map to construct the scene representations as well as the $Size$ of constructed maps that reflect the memory usage, the time $T_e$ for traversability evaluation and obtaining the navigation cost maps, the time $T_s$ for searching the initial paths, the number of nodes $N_s$ that the graph traversal algorithm visits, the time $T_o$ for trajectory optimization, and the total time $T_{all}$ to solve the navigation problem.
        The operations on GPU are synchronized for accurate measurement of the computation time. 
        The time of Wang's method to build 3D occupancy grids is excluded from $T_p$ and $T_{all}$ for fair comparison as its CPU implementation is inefficient.
        For trajectory evaluation in \textit{plaza} and \textit{hills}, we compare the average time for path searching $\overline{T_s}$ and optimization $\overline{T_o}$, the average trajectory length $\overline{L_t}$ and curvature $\overline{C_t}$, and the success rate $S_t$ of finding feasible trajectories.
        All the approaches are evaluated on a desktop with an Intel i9-12900KF @ 3.2GHz CPU and NVIDIA RTX 3080 Ti GPU.
        In addition, we deploy our approach on an NVIDIA Jetson AGX Orin (MAXN mode) to further demonstrate the high efficiency of our approach even on a mobile computation device.

\begin{figure}[t]
\setlength{\abovecaptionskip}{0pt}
\centering
\includegraphics[width=2.8in]{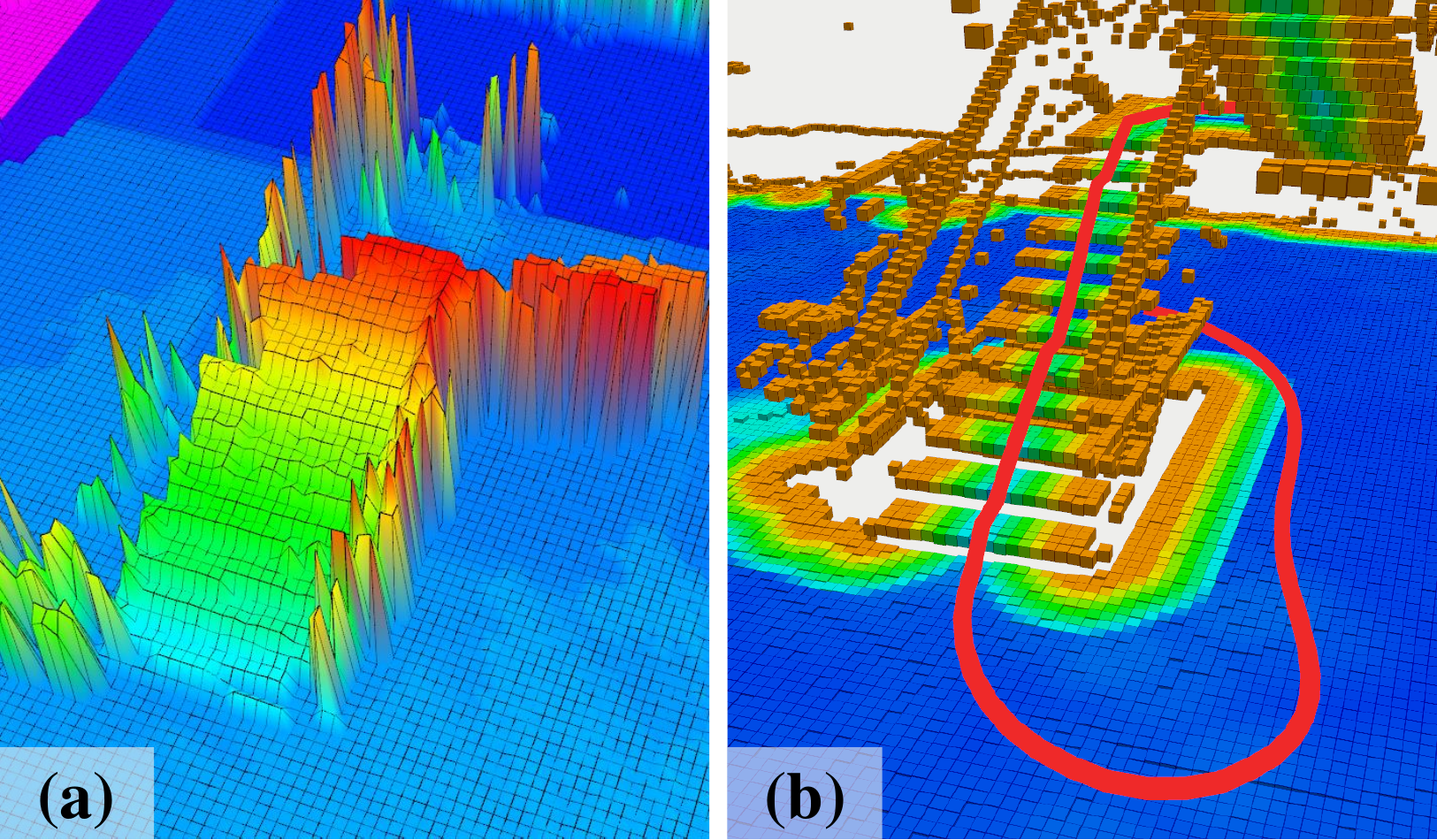}
\caption{
(a): Yang's method \cite{yang2021RTOptimal} using traditional elevation maps \cite{elevationCupy} only evaluates a single terrain layer at a time and fails to plan in \textit{stairs}. 
(b): Our approach extends elevation maps to represent 3D planning spaces and generates trajectory from the place below the stairs to the platform above.
}
\label{fig:emc}
\vspace{-1.2em}
\end{figure}

\renewcommand{\arraystretch}{1.0}
\renewcommand\tabcolsep{3.5pt}
\vspace{-0.8em}
\begin{table}[h]
\caption{Scene Evaluation and Trajectory Generation Capabilities}
\centering
\begin{threeparttable}
\begin{tabular}{l|cccc|c}
    \toprule
    Methods & \scriptsize \makecell[c]{Planning in \\ 3D Spaces} & \scriptsize \makecell[c]{Body Height \\ Adaptation} & \scriptsize \makecell[c]{Planning on \\ the Stairs} & \scriptsize \makecell[c]{Motion \\ Capabilities \\ Awareness} & \scriptsize \makecell[c]{Trajectory \\ with \\ Velocities} \\ 
    \midrule 
    Yang's \cite{yang2021RTOptimal} & \color{Red}\ding{56}   & \color{Red}\ding{56}   & \color{Green}\ding{52} & \color{Green}\ding{52} & \color{Red}\ding{56} \\
    Liu's \cite{roboticOnline}      & \color{Green}\ding{52} & \color{Red}\ding{56}   & \color{Green}\ding{52} & \color{Red}\ding{56}   & \color{Red}\ding{56}   \\
    Pütz's \cite{spvfMesh}          & \color{Green}\ding{52} & \color{Red}\ding{56}   & \color{Red}\ding{56}   & \color{Green}\ding{52} & \color{Red}\ding{56}   \\
    Wang's \cite{beyond2D}          & \color{Green}\ding{52} & \color{Green}\ding{52} & \color{Red}\ding{56}   & \color{Green}\ding{52} & \color{Green}\ding{52} \\
    Ours                            & \color{Green}\ding{52} & \color{Green}\ding{52} & \color{Green}\ding{52} & \color{Green}\ding{52} & \color{Green}\ding{52} \\
    \bottomrule
\end{tabular}
\footnotesize
\end{threeparttable}
\label{tab:cap}
\vspace{-0.8em}
\end{table}
\renewcommand{\arraystretch}{1.0}

\renewcommand{\arraystretch}{0.85}
\renewcommand\tabcolsep{5.2pt}
\begin{table*}[t]
\caption{Evaluation of Computation and Memory Efficiency in Map Construction, Scene Evaluation, and Trajectory Generation}
\centering
\begin{threeparttable}
\begin{tabular}{lc|cc|c|ccc|c}
    \toprule
    \multirow{2}{*}{\textit{Scenarios}} 
        & \multirow{2}{*}{\textit{Approaches}}
            & \multicolumn{2}{c|}{Map Construction} & Scene Evaluation & \multicolumn{3}{c|}{Trajectory Generation} & Overall \\ 
        \cmidrule{3-9}
        &   & $T_p[\unit{\ms}]$ $\downarrow$ & $Size[\unit{\mega\byte}]$ $\downarrow$ & $T_e[\unit{\ms}]$ $\downarrow$ & $T_s[\unit{\ms}]$ $\downarrow$ & $N_s$ $\downarrow$ & $T_o[\unit{\ms}]$ $\downarrow$ & $T_{all}[\unit{\ms}]$ $\downarrow$ \\ 
    \midrule
    \multirow{6}{*}{\textit{Factory}} 
        & Liu's \cite{roboticOnline} & $-$          & \num{12.31}  & \num{178.28e3} & \num{189.50}  & $135,290$      & $-$           & \num{178.75e3} \\
        & Pütz's \cite{spvfMesh}     & \num{3.44e3} & \num{28.84}  & \num{13.36e3}  & \num{100.94}  & $\bm{80,333}$  & $-$           & \num{16.91e3}  \\
        & Wang's \cite{beyond2D}     & \num{2.22e3} & \num{34.27}  & \num{24.54e3}  & \num{1.37e3}  & $3,892,807$    & \num{953.46}  & \num{29.08e3}  \\
        & Ours (PC-CPU only)    & \num{630.76} & $\bm{4.57}$  & \num{785.68}   & $\bm{25.66}$  & $133,648$      & $\bm{475.64}$ & \num{1.92e3}   \\
        & Ours (PC)                  & $\bm{2.39}$  & $\bm{4.57}$  & $\bm{2.41}$    & $\bm{25.66}$  & $133,648$      & $\bm{475.64}$ & $\bm{542.55}$  \\
        & Ours (Orin)                & \num{20.06}  & $\bm{4.57}$  & \num{12.53}    & \num{75.56}   & $133,648$      & \num{1.36e3}  & \num{1.56e3}   \\
    \midrule
    \multirow{6}{*}{\textit{Building}} 
        & Liu's \cite{roboticOnline} & $-$          & \num{7.40}  & \num{116.47e3} & \num{765.54}  & $208,973$      & $-$           & \num{117.51e3} \\
        & Pütz's \cite{spvfMesh}     & \num{3.41e3} & \num{17.10} & \num{11.36e3}  & $-$           & $-$            & $-$           & $-$            \\
        & Wang's \cite{beyond2D}     & \num{1.91e3} & \num{28.16} & \num{11.53e3}  & $-$           & $-$            & $-$           & $-$            \\
        & Ours (PC-CPU only)    & \num{290.98} & $\bm{3.17}$ & \num{234.95}   & $\bm{7.33}$   & $\bm{39,760}$  & $\bm{359.39}$ & \num{892.65}   \\
        & Ours (PC)                  & $\bm{3.10}$  & $\bm{3.17}$ & $\bm{3.42}$    & $\bm{7.33}$   & $\bm{39,760}$  & $\bm{359.39}$ & $\bm{398.14}$  \\
        & Ours (Orin)                & \num{23.05}  & $\bm{3.17}$ & \num{17.69}    & \num{17.89}   & $\bm{39,760}$  & \num{989.66}  & \num{1.11e3}   \\
    \midrule
    \multirow{6}{*}{\textit{Forest}} 
        & Liu's \cite{roboticOnline} & $-$          & \num{6.94}   & \num{93.86e3}  & \num{66.24}   & $\bm{29,234}$  & $-$           & \num{94.18e3}  \\
        & Pütz's \cite{spvfMesh}     & \num{4.52e3} & \num{17.94}  & \num{12.88e3}  & \num{68.27}   & $57,162$       & $-$           & \num{17.47e3}  \\
        & Wang's \cite{beyond2D}     & \num{1.79e3} & \num{44.80}  & \num{24.67e3}  & \num{917.61}  & $3,014,581$    & $\bm{42.95}$  & \num{27.42e3}  \\
        & Ours (PC-CPU only)    & \num{293.43} & $\bm{6.40}$  & \num{460.51}   & $\bm{19.56}$  & $85,450$       & \num{65.43}   & \num{838.93}  \\
        & Ours (PC)                  & $\bm{1.42}$  & $\bm{6.40}$  & $\bm{3.02}$    & $\bm{19.56}$  & $85,450$       & \num{65.43}   & $\bm{109.99}$  \\
        & Ours (Orin)                & \num{12.50}  & $\bm{6.40}$  & \num{19.88}    & \num{60.50}   & $85,450$       & \num{175.75}  & \num{317.40}   \\
    \midrule
    \multirow{6}{*}{\textit{Overpass}} 
        & Liu's \cite{roboticOnline} & $-$           & $\bm{16.35}$  & \num{1.19e6}   & \num{820.49}  & $596,386$      & $-$           & \num{1.19e6}   \\
        & Pütz's \cite{spvfMesh}     & \num{18.30e3} & \num{67.73}   & \num{45.21e3}  & \num{840.81}  & $399,943$      & $-$           & \num{64.35e3}  \\
        & Wang's \cite{beyond2D}     & \num{4.56e3}  & \num{220.88}  & \num{97.64e3}  & \num{19.81e3} & $55,489,651$   & \num{5.24e3}  & \num{127.26e3} \\
        & Ours (PC-CPU only)    & \num{1.49e3}  & \num{26.65}   & \num{3.30e3}   & $\bm{37.82}$  & $\bm{224,651}$ & $\bm{2.91\times10^3}$ & \num{7.74e3}  \\
        & Ours (PC)                  & $\bm{18.11}$  & \num{26.65}   & $\bm{14.35}$   & $\bm{37.82}$  & $\bm{224,651}$ & $\bm{2.91\times10^3}$ & $\bm{3.17\times10^3}$  \\
        & Ours (Orin)                & \num{123.55}  & \num{26.65}   & \num{81.78}    & \num{92.59}   & $\bm{224,651}$ & \num{7.55e3}  & \num{8.19e3}   \\
    \midrule \midrule
    \multirow{6}{*}{\textit{Stairs}} 
        & Liu's \cite{roboticOnline} & $-$          & \num{1.34}  & \num{10.46e3} & \num{67.68} & $43,302$      & $-$          & \num{10.75e3} \\
        & Pütz's \cite{spvfMesh}     & \num{3.19e3} & \num{3.69}  & \num{3.36e3}  & $-$         & $-$           & $-$          & $-$           \\
        & Wang's \cite{beyond2D}     & \num{346.40} & \num{5.49}  & \num{2.58e3}  & $-$         & $-$           & $-$          & $-$           \\
        & Ours (PC-CPU only)    & \num{55.41}  & $\bm{0.47}$ & \num{58.10}   & $\bm{2.06}$ & $\bm{12,426}$ & $\bm{54.29}$ & \num{169.86}  \\
        & Ours (PC)                  & $\bm{0.42}$  & $\bm{0.47}$ & $\bm{0.20}$   & $\bm{2.06}$ & $\bm{12,426}$ & $\bm{54.29}$ & $\bm{61.23}$  \\
        & Ours (Orin)                & \num{3.23}   & $\bm{0.47}$ & \num{1.36}    & \num{6.03}  & $\bm{12,426}$ & \num{132.19} & \num{161.50}  \\
    \midrule
    \multirow{6}{*}{\textit{Boxes}} 
        & Liu's \cite{roboticOnline} & $-$          & \num{2.75}  & \num{22.56e3} & \num{19.00} & $5,935$       & $-$          & \num{22.82e3} \\
        & Pütz's \cite{spvfMesh}     & \num{1.62e3} & \num{5.10}  & \num{1.61e3}  & \num{3.91}  & $\bm{2,875}$  & $-$          & \num{3.23e3}  \\
        & Wang's \cite{beyond2D}     & \num{935.63} & \num{7.74}  & \num{3.55e3}  & \num{90.94} & $245,412$     & \num{16.00}  & \num{4.60e3}  \\
        & Ours (PC-CPU only)    & \num{107.55} & $\bm{1.16}$ & \num{18.70}   & $\bm{1.64}$ & $7,794$       & $\bm{14.62}$ & \num{142.51}  \\
        & Ours (PC)                  & $\bm{0.37}$  & $\bm{1.16}$ & $\bm{0.95}$   & $\bm{1.64}$ & $7,794$       & $\bm{14.62}$ & $\bm{24.34}$  \\
        & Ours (Orin)                & \num{2.93}   & $\bm{1.16}$ & \num{5.52}    & \num{4.53}  & $7,794$       & \num{41.65}  & \num{75.72}   \\
    \bottomrule
\end{tabular}
\footnotesize
\begin{tablenotes}
    \item Evaluation results of the navigation approaches in simulated (row 1 to 4) and real-world (row 5, 6) scenarios. 
    $T_p$: point cloud pre-processing time, $Size$: memory usage of the constructed scene representation, $T_e$: traversability estimation time, $T_s$: path searching time, $N_s$: visited graph nodes for planning, $T_o$: trajectory optimization time, $T_{all}$: total navigation time.
\end{tablenotes}
\end{threeparttable}
\label{tab:simu_real}
\vspace{-1.0em}
\end{table*}
\renewcommand{\arraystretch}{1.0}

        \subsection{Navigation Capabilities}
        The first four items in Table \ref{tab:cap} compare the scene evaluation capabilities of different methods.
        Although Yang's method \cite{yang2021RTOptimal} plans with motion capabilities awareness of robots on stairs and other complex terrains, it fails to plan in multi-layer 3D spaces as the traditional elevation map \cite{elevationCupy} only recognizes a single terrain layer at a time (Fig. \ref{fig:emc}).
        Other approaches can navigate in 3D multi-layer scenarios.
        However, only Wang's and our approach can adjust the robot's body height along the $z$-axis for active adaptation to narrow environments.
        Both Pütz's and Wang's methods are unsuitable for planning on the stairs for legged robots, for they consider the vertical surfaces of the steps as untraversable.
        Although Liu's approach plans on the stairs, it fails to capture the motion capabilities of different robots and may generate infeasible paths.
        The last item in Table \ref{tab:cap} compares the trajectory generation capability.
        Only Wang's and our approach generate trajectories with velocity information, as it's normally easier to optimize paths on gird maps.
        While Pütz \textit{et al.} \cite{spvfMesh} also present the \ac{CVP} to generate smooth paths, it does not plan the robot's velocities.
        Our approach provides smooth trajectories considering the robot's capabilities of locomotion and body height adjustment in complex environments, presenting strong potential in a wide range of navigation applications.

\renewcommand{\arraystretch}{0.85}
\renewcommand\tabcolsep{13.5pt}
\begin{table*}[t]
\caption{Evaluation of Trajectory Generation Performance in Run-Time and Quality}
\centering
\begin{threeparttable}
\begin{tabular}{lc|ccccc}
    \toprule
    \textit{Scenarios} & \textit{Approaches}
    & $\overline{T_s}[\unit{\ms}]$ $\downarrow$ & $\overline{T_o}[\unit{\ms}]$ $\downarrow$ & $\overline{L_t}[\unit{\m}]$ $\downarrow$ & $\overline{C_t}[\unit{\per\m}]$ $\downarrow$ & $S_t$ $\uparrow$ \\ 
    \midrule
    \multirow{4}{*}{\textit{Plaza}} 
        & Liu's \cite{roboticOnline} & \num{358.17} $\pm$ \num{25.78}  & $-$                            & \num{28.91}  $\pm$ \num{0.71} & \num{0.353}  $\pm$ \num{0.645} & \num{0.12}  \\
        & Pütz's \cite{spvfMesh}     & \num{44.16}  $\pm$ \num{4.43}   & $-$                            & \num{30.69}  $\pm$ \num{1.53} & \num{0.164}  $\pm$ \num{0.018} & \num{0.92}  \\
        & Wang's \cite{beyond2D}     & \num{1.19e3} $\pm$ \num{335.25} & $\bm{33.83}$ $\pm$ \num{19.19} & \num{47.13}  $\pm$ \num{9.59} & \num{0.037}  $\pm$ \num{0.010} & \num{0.94}  \\
        & Ours (PC)                  & $\bm{13.93}$ $\pm$ \num{3.93}   & \num{52.57}  $\pm$ \num{25.99} & $\bm{28.32}$ $\pm$ \num{1.55} & $\bm{0.018}$ $\pm$ \num{0.007} & $\bm{1.00}$ \\
    \midrule
    \multirow{4}{*}{\textit{Hills}} 
        & Liu's \cite{roboticOnline} & \num{219.31} $\pm$ \num{9.19}   & $-$                             & \num{29.65}  $\pm$ \num{0.57}  & \num{0.149}  $\pm$ \num{0.015} & \num{0.28}  \\
        & Pütz's \cite{spvfMesh}     & \num{62.42}  $\pm$ \num{3.36}   & $-$                             & \num{28.79}  $\pm$ \num{0.30}  & \num{0.130}  $\pm$ \num{0.031} & \num{0.80}  \\
        & Wang's \cite{beyond2D}     & \num{1.05e3} $\pm$ \num{528.27} & \num{33.58}  $\pm$ \num{30.33}  & \num{40.53}  $\pm$ \num{13.73} & $\bm{0.011}$ $\pm$ \num{0.010} & \num{0.86}  \\
        & Ours (PC)                  & $\bm{14.23}$ $\pm$ \num{4.39}   & $\bm{32.89}$ $\pm$ \num{22.87}  & $\bm{27.79}$ $\pm$ \num{0.44}  & $\bm{0.011}$ $\pm$ \num{0.006} & $\bm{0.92}$ \\
    \bottomrule
\end{tabular}
\footnotesize
\begin{tablenotes}
    \item Trajectory evaluation results in simulated \textit{plaza} and \textit{hills} scenarios (mean and standard deviation). 
    $\overline{T_s}$: average path searching time, $\overline{T_o}$: average trajectory optimization time, $\overline{L_t}$: average trajectory length, $\overline{C_t}$: average trajectory curvature, $S_t$: success rate of finding feasible trajectories in 50 attempts.
\end{tablenotes}
\end{threeparttable}
\label{tab:traj50}
\vspace{-1.0em}
\end{table*}
\renewcommand{\arraystretch}{1.0}

        \subsection{Simulation Results}
        The first four rows of Table \ref{tab:simu_real} present the evaluation results in simulation scenarios. 
        For \textit{building}, both Pütz's and Wang's methods can not recognize the traversable stairs, thus resulting in the failure of planning.
        As our map construction and scene evaluation steps are able to be accelerated through parallel computation, the processing speed is significantly improved by 2 to 3 orders of magnitude with less $T_p$ and $T_e$.
        The constructed maps also maintain high efficiency in memory usage with smaller $Sizes$ using our proposed representation approach.
        Although Liu's approach also runs on GPU, the tensor voting process is still quite time-consuming.
        Also, Wang's method takes a long time to perform plane fitting, resulting in large $T_e$.
        In addition, our approach reduces the burden of path searching with less time $T_s$, which benefits the tasks with re-planning requirements.
        Wang's method directly plans on dense 3D voxels, which introduces unnecessarily large numbers of node visits. 
        Both Liu's and Pütz's approaches have small $N_s$, as they might generate navigation graphs sparser than grid maps by evaluating points or meshes.
        However, they take a longer time to query the neighboring nodes and finish a single visit compared with searching on grid maps. 
        The total time $T_{all}$ of our approach which already includes the time for data exchange between computation devices is still small.
        Our approach generates trajectories in a short time even on a mobile device. 
        The computation time is still rational in large-scale scenarios like \textit{overpass}.

        Fig. \ref{fig:tomo_sim} presents the 3D models (row 1), the point clouds (row 2), and the traversability estimation results of our approach (row 3) together with our generated trajectories.
        Our approach provides feasible, smooth, and executable trajectories in both indoor and outdoor multi-layer environments with structured or irregular terrains.
        The trajectories in \textit{factory}, \textit{forest}, and \textit{overpass} are successfully validated on the Pioneer 3-DX robot in physical simulation (please see the attached video for more detail on the simulation results).

\begin{figure}[t]
\setlength{\abovecaptionskip}{0pt}
\centering
\includegraphics[width=2.5in]{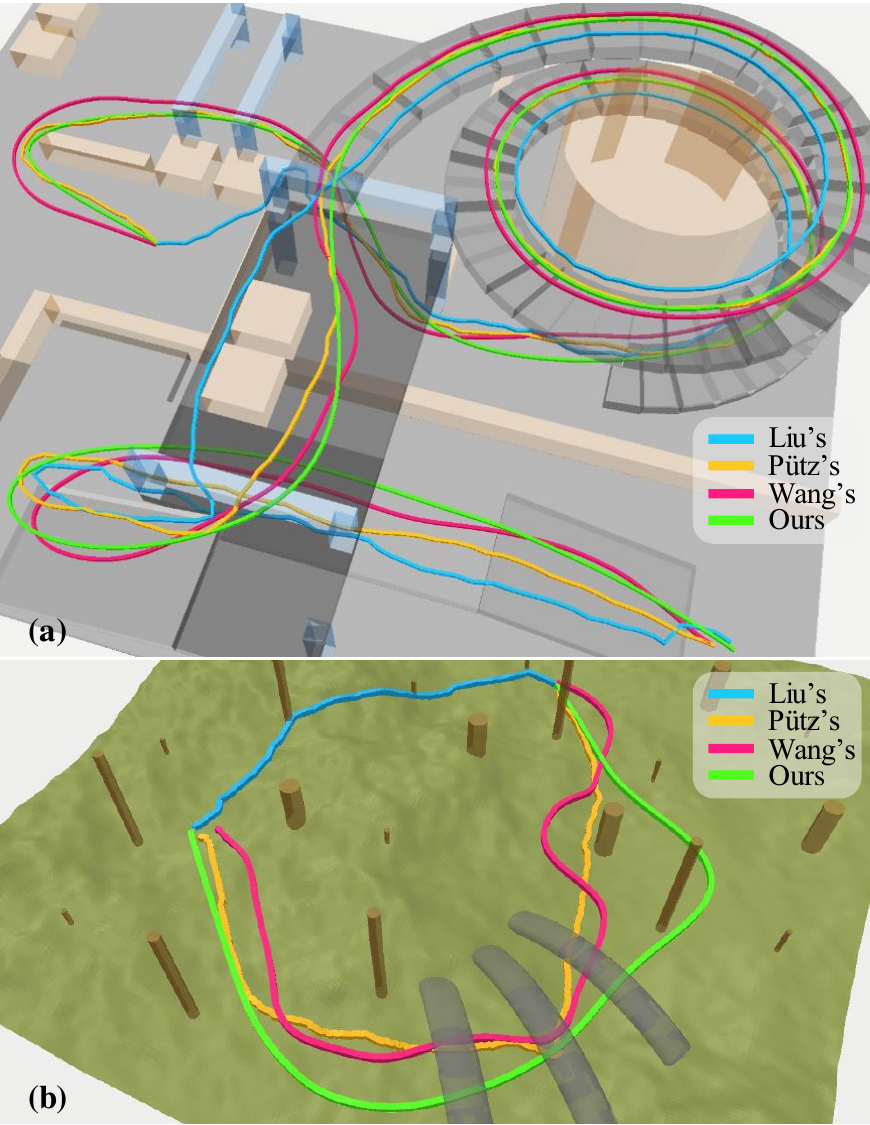}
\caption{
Planning results of the evaluated approaches in \textit{factory} (a) and \textit{forest} (b).
Green: our approach, blue: Liu's method, yellow: Pütz's method, red: Wang's method.
Liu's method provides infeasible solutions as it fails to consider the robot's shape and motion capabilities.
Pütz's method provides paths without velocity information.
Wang's trajectory flies down from the side of the slope in the bottom left corner of (a), resulting in irrational behavior.
}
\label{fig:traj}
\vspace{-1.0em}
\end{figure}

        Fig. \ref{fig:traj} compares the planning results of different approaches in \textit{factory} (a) and \textit{forest} (b). 
        Without the awareness of the robot's motion capabilities, Liu's approach (blue) generates infeasible paths.
        It fails to consider the body size of the robot in \textit{factory} and the path collides with the pillars in \textit{forest} as it shows a strong preference for following the geodesic direction.
        Pütz's method (yellow) plans feasible paths without providing the velocity information.
        Both Wang's (red) and our approach (green) generate smooth trajectories.
        However, Wang's method may fail to capture detailed terrain structures using voxels and provide sub-optimal solutions.
        For example, in \textit{factory}, the red trajectory takes an early turn and flies down the first slope from the side (bottom left corner of Fig. \ref{fig:traj} (a)).

        Table \ref{tab:traj50} evaluates the trajectory generation performance. 
        Without the awareness of robots' motion capabilities, Liu's method \cite{roboticOnline} frequently moves through untraversable obstacles or narrow passages.
        Wang's method \cite{beyond2D} using rough voxels for scene evaluation fails to capture detailed terrain information and plans with unnecessary detours, resulting in long trajectory length.
        Our approach provides high-quality smooth trajectories in a short time with lower trajectory length $\overline{L_t}$ and curvature $\overline{C_t}$ on average in both scenarios.
        By analyzing continuous ground elevations, our approach better understands the terrain conditions and achieves higher success rates.

\begin{figure}[h]
\setlength{\abovecaptionskip}{0pt}
\centering
\includegraphics[width=3.0in]{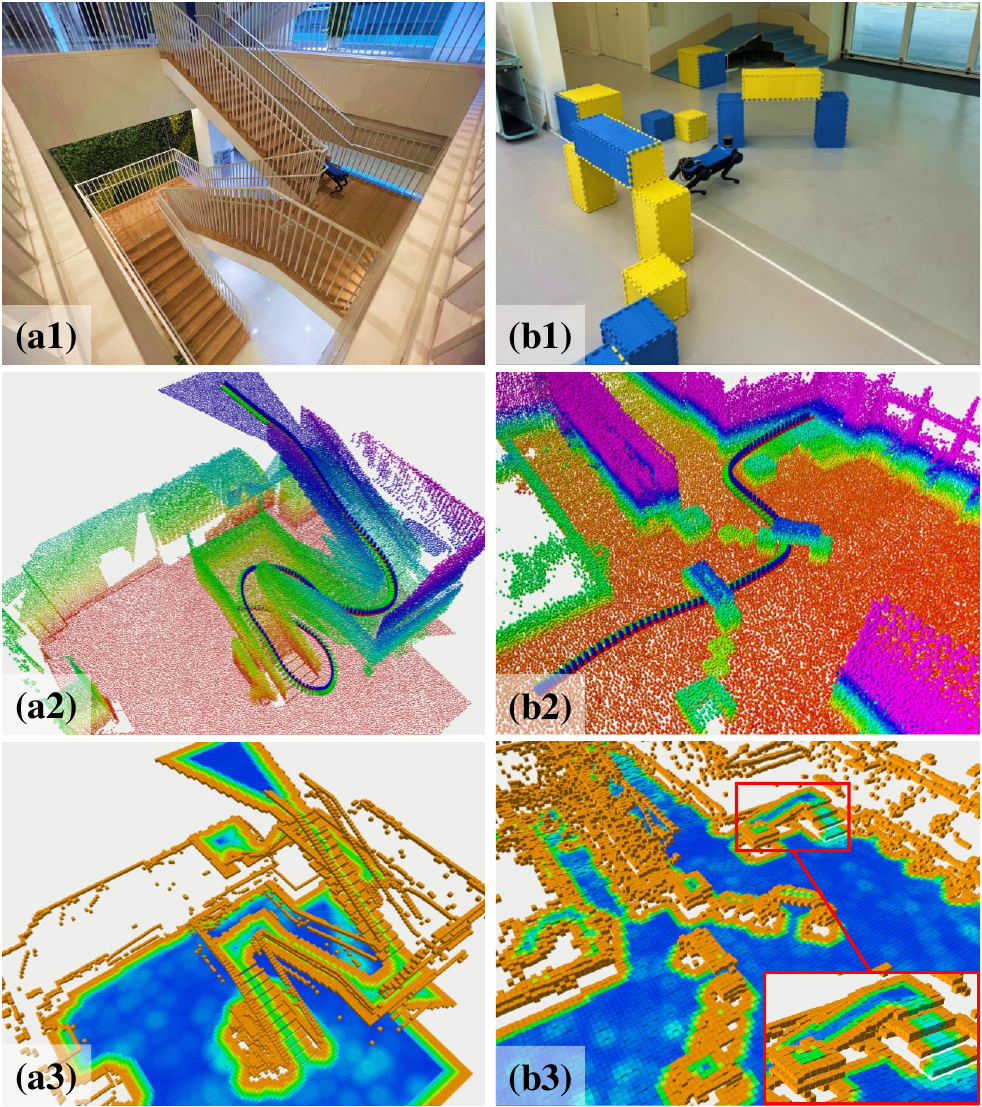}
\caption{
The real-world experiment scenarios and the navigation results. 
The first row shows the photos of environments: \textit{stairs} (a1) and \textit{boxes} (b1). 
Row 2 presents the corresponding point cloud maps and row 3 shows the integrated travel cost maps generated by our approach.
Our quadrupedal robot successfully finished these navigation tasks in real-world experiments following the generated trajectories. 
In \textit{boxes}, the robot prefers to move up to the platform through the gentle slope as the slope has lower travel costs and is safer than the steep stairs.
}
\label{fig:tomo_real}
\vspace{-1.0em}
\end{figure}

        \subsection{Real-world Performance}
        The last two rows of Table \ref{tab:simu_real} show the computation time of the evaluated approaches in real-world experiments.
        Our approach still generates feasible trajectories on the point cloud maps and outperforms the baseline approaches using other scene representations in efficiency.
        It performs scene evaluation and initial path planning in only several milliseconds and the total navigation speed is more than 2 orders of magnitude faster than existing approaches.
        
        Fig. \ref{fig:tomo_real} shows the photos (row 1), point cloud maps with the generated trajectories (row 2), and the estimated travel costs (row 3) of the real-world scenarios.
        In \textit{stairs}, our approach successfully navigates the quadrupedal robot to reach the second floor through the winding stairs in this multi-layer structure.
        In addition, by capturing the terrain features and rationally designing the travel cost functions, our framework navigates the robot with locomotion capabilities awareness. 
        The robot can automatically adjust the body height to walk through the narrow arches in \textit{boxes}.
        Also, it chooses to move up from the slope rather than the stairs beside to reach the target as walking on the steep stairs is more risky and the travel costs on the stairs are higher.
        More details of the real-world experiments are also presented in our video.

\section{Discussion}
In addition to its strong performance in the experiments, the proposed navigation framework is also extensible to be applied in more complex navigation tasks.
Our scene representation is compatible with a wide range of traditional or learning-based methods on elevation maps to further enhance the performance of map construction, scene evaluation, and path planning. 
Semantic information could also be concatenated to the tomogram slices to provide additional environment features beyond the geometric structures for more robust navigation behavior.
By understanding the 3D environment as 2.5D layers, our framework achieves high simplicity and computation efficiency, which also presents its strong potential for online navigation and exploration tasks.

There are some limitations of the presented framework.
The current approach requires a relatively dense and high-quality point cloud map of the scene.
It may generate sub-optimal trajectories or fail to find a solution if the point cloud is too sparse or contains too much noise and dynamic objects. 
Although existing methods can link faraway nodes across the missing areas for path planning \cite{roboticOnline,spvfMesh} or fill in the blanks with neighboring elevations \cite{elevationCupy} for dense mapping, the lack of concrete terrain information in unobserved regions may lead to dangers in downstream tasks \cite{yang2023neural}.
Learning-based approaches could be introduced to recover the structure of noisy or occluded regions better with the awareness of reconstruction uncertainty (as done in \cite{yang2023neural}) and remove dynamic points to reduce human burden in pre-processing point clouds.

\section{Conclusion}
This paper presented a highly efficient and extensible global navigation framework for ground robots in complex multi-layer structures. 
We introduced a novel scene representation to analyze the point cloud map from a tomographic view.
The resulting tomogram slices extend traditional elevation maps to represent multi-layer 3D structures while maintaining their simplicity in mapping and processing.
Both terrain conditions and spatial structures are evaluated with the awareness of the robot's locomotion and height adjustment capabilities. 
The map construction and the scene evaluation stages can be accelerated through parallel computation to reduce the processing time.
In addition, our scene representation approach reduces the burden of path search.
Our trajectory generation module efficiently provides cost-optimal 3D trajectories and supports active body height adaptation to the narrow environments.
The proposed framework is evaluated in both simulated and real-world experiments, demonstrating its highly efficient navigation performance in various complicated 3D environments.

\section*{Acknowledgment}
We would like to thank Ren Xin, Peng Yun, and Xiangcheng Hu for their valuable suggestions.

\bibliography{reference} 
\bibliographystyle{ieeetr}

\vspace{-4em}

\begin{IEEEbiography}[{\includegraphics[width=0.9in,height=1.25in,clip,keepaspectratio]{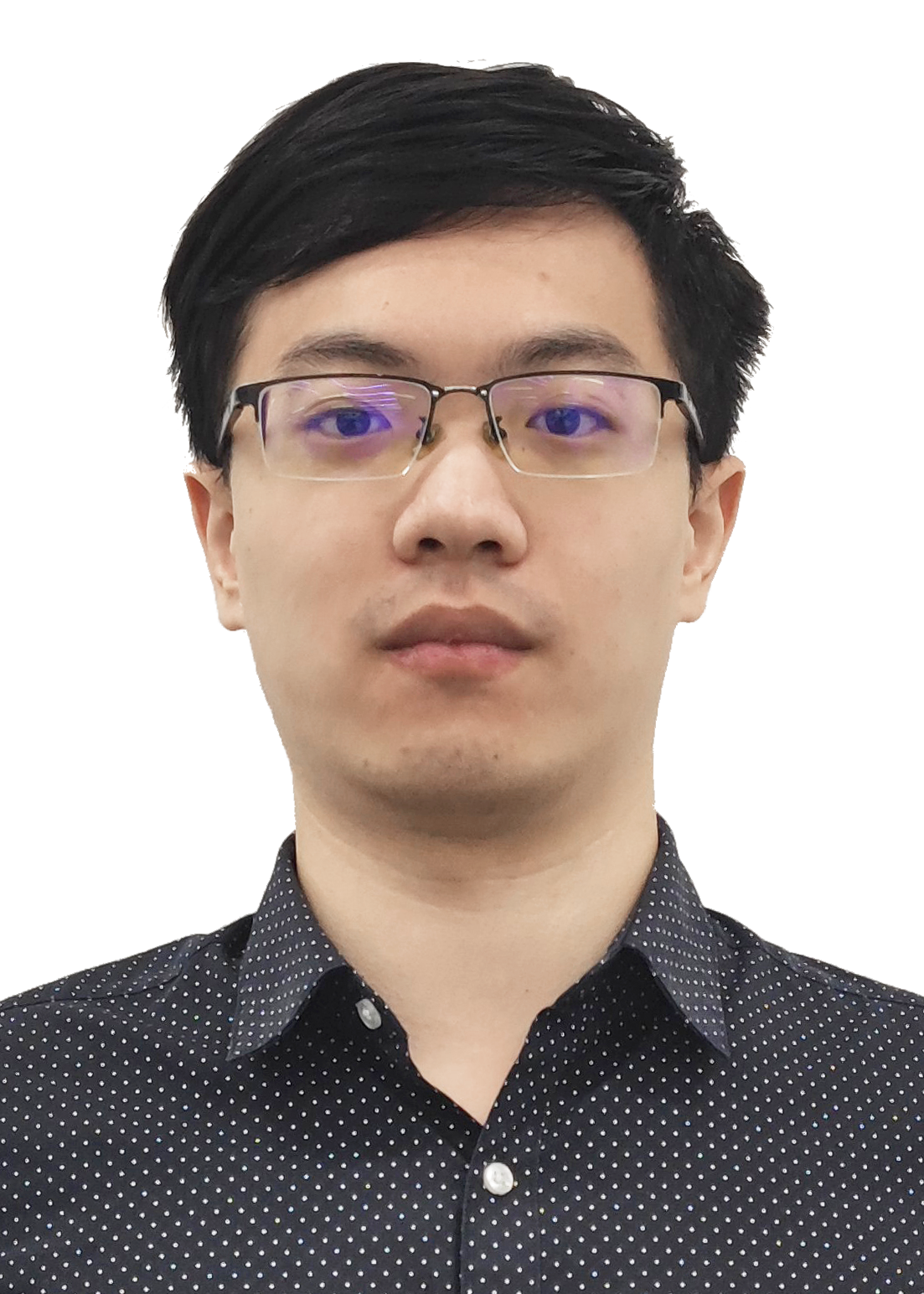}}]{Bowen Yang}
    received the B.Eng. degree in the Hong Kong Polytechnic University, HKSAR, China, in 2019. 
    He is currently pursuing the Ph.D. degree in the Department of Electronic and Computer Engineering, the Hong Kong University of Science and Technology, HKSAR, China, supervised by Prof. Ming Liu.
    His research mainly focuses on scene understanding and path planning for autonomous robots.
\end{IEEEbiography}
\vspace{-4em}

\begin{IEEEbiography}[{\includegraphics[width=0.9in,height=1.25in,clip,keepaspectratio]{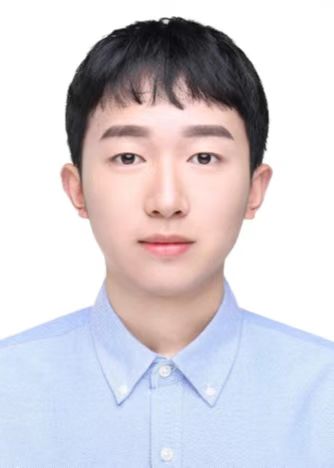}}]{Jie Cheng}
     received the B.S. degree from Huazhong University of Science and Technology, Wuhan, China, in 2019. 
     He is currently pursuing the Ph.D. degree in the Department of Electronic and Computer Engineering, the Hong Kong University of Science and Technology, HKSAR, China, supervised by Prof. Ming Liu.
     His research mainly focuses on motion planning and motion forecasting for robots and autonomous vehicles.
\end{IEEEbiography}
\vspace{-4em}

\begin{IEEEbiography}[{\includegraphics[width=0.9in,height=1.25in,clip,keepaspectratio]{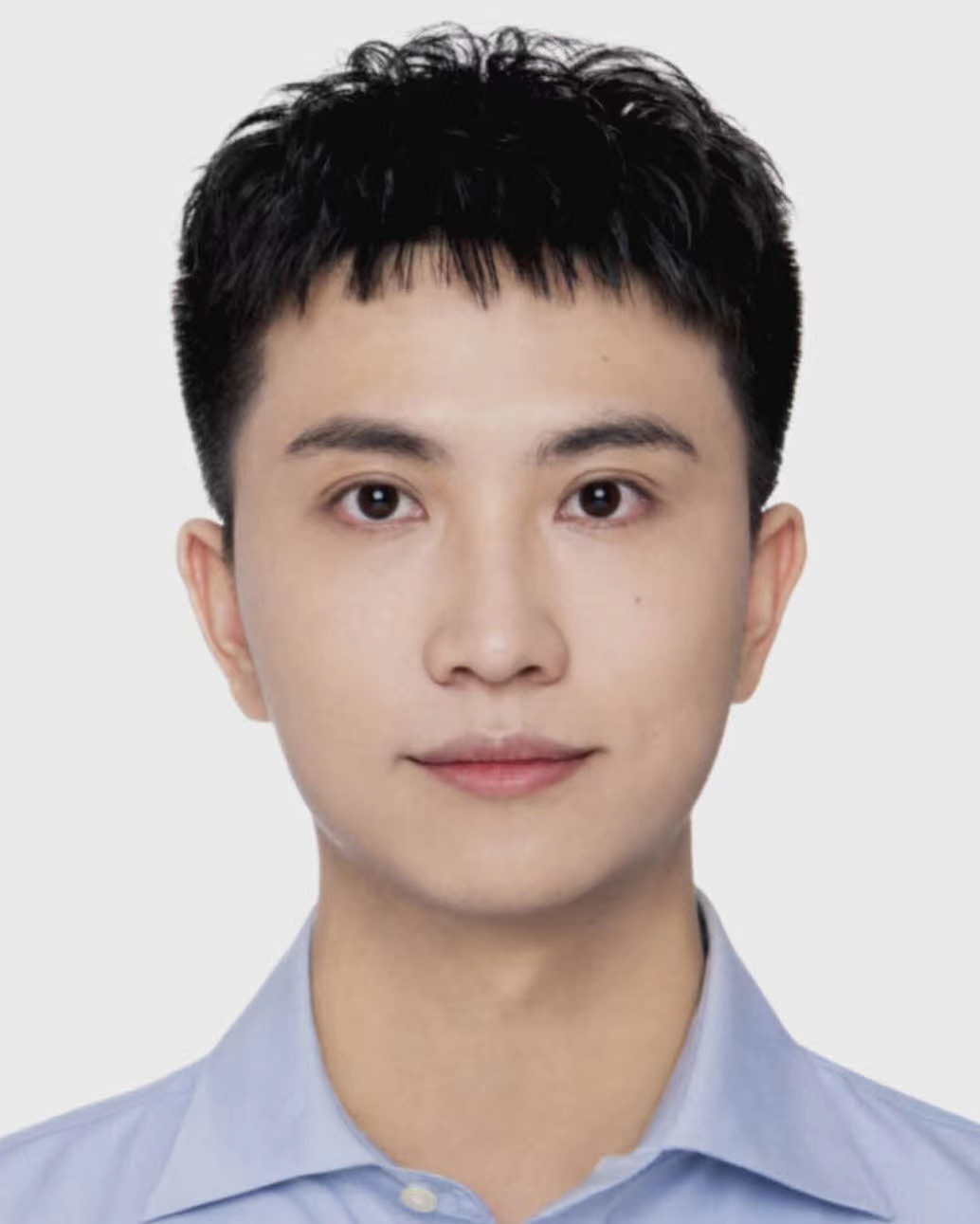}}]{Bohuan Xue}
    received the B.Eng. degree in computer science and technology from College of Mobile Telecommunications, Chongqing University of Posts and and Telecom, Chongqing, China, in 2018. 
    He is currently working toward the Ph.D. degree in electrical engineering with the Department of Computer Science and Engineering, the Hong Kong University of Science and Technology, HKSAR, China. 
    His research interests include SLAM, computer vision, and 3D reconstruction.
\end{IEEEbiography}
\vspace{-4em}

\begin{IEEEbiography}[{\includegraphics[width=0.9in,height=1.25in,clip,keepaspectratio]{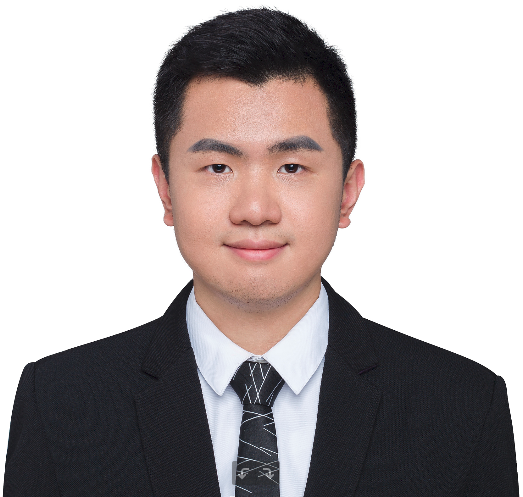}}]{Jianhao Jiao}
    received the B.Eng. degree in instrument science from Zhejiang University, Hangzhou, China, in 2017, and the Ph.D. from the Department of Electronic and Computer Engineering, the Hong Kong University of Science and Technology, HKSAR, China, in 2021, supervised by Prof. Ming Liu.
    He is now a research associate at the same university.
    His research interests include state estimation, SLAM, dense mapping, sensor fusion, and computer vision.
\end{IEEEbiography}
\vspace{-4em}

\begin{IEEEbiography}[{\includegraphics[width=0.9in,height=1.25in,clip,keepaspectratio]{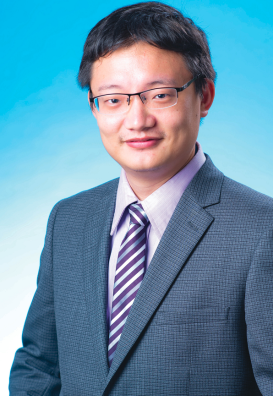}}]{Ming Liu}
    received the Ph.D. degree from the Department of Mechanical and Process Engineering, ETH Zurich, Switzerland, in 2013, supervised by Prof. Roland Siegwart.
    He is currently with the Robotics and Autonomous Systsms, The Hong Kong University of Science and Technology (Guangzhou), the director of Intelligent Autonomous Driving Center, as an Associate Professor.
    Prof. Liu is currently an Associate Editor for IEEE Robotics and Automation Letters, IET Cyber-Systems and Robotics, International Journal of Robotics and Automation.
    His research interests include dynamic environment modeling, deep-learning for robotics, 3D mapping, machine learning, and visual control.
\end{IEEEbiography}

\end{document}